\documentclass[10pt,twocolumn,letterpaper]{article}

\usepackage[pagenumbers]{cvpr} 

\usepackage{graphicx}
\usepackage{amsmath}
\usepackage{amssymb}
\usepackage{booktabs}
\usepackage{xspace}
\usepackage{multirow}
\usepackage{algcompatible}

\usepackage{hyperref}       
\usepackage{url}            
\usepackage{amsfonts}       
\usepackage{nicefrac}       
\usepackage{microtype}      
\usepackage{xcolor}         
\usepackage{algorithm} 
\usepackage{algpseudocode}
\usepackage{hyperref}
\usepackage{subcaption}
\usepackage[capitalize]{cleveref}
\hypersetup{colorlinks,linkcolor={blue},citecolor={blue},urlcolor={red}}  
\makeatletter
\DeclareRobustCommand\onedot{\futurelet\@let@token\@onedot}
\def\@onedot{\ifx\@let@token.\else.\null\fi\xspace}

\def\eg{\emph{e.g}\onedot} 
\def\ie{\emph{i.e}\onedot}

\makeatother

\usepackage[capitalize]{cleveref}
\crefname{section}{Sec.}{Secs.}
\Crefname{section}{Section}{Sections}
\Crefname{table}{Table}{Tables}
\crefname{table}{Tab.}{Tabs.}

\begin{document}

\title{StraIT: Non-autoregressive Generation with Stratified Image Transformer}

\author{Shengju Qian$^{1,3}$\thanks{This work was done during an internship at Google Research. Correspondence to: Han Zhang \texttt{<zhanghan@google.com>}, Shengju Qian \texttt{<sjqian@cse.cuhk.edu.hk>}. Code and models will be released.}
\quad Huiwen Chang$^1$
\quad Yuanzhen Li$^1$
\quad Zizhao Zhang$^2$ \\
Jiaya Jia$^3$
\quad  Han Zhang$^1$ \\
$^1$Google Research \quad $^2$Google Cloud AI \quad $^3$CUHK
}
\maketitle

\begin{abstract}
    We propose Stratified Image Transformer~(StraIT), a pure non-autoregressive~(NAR) generative model that demonstrates superiority in high quality image synthesis over existing autoregressive~(AR) and diffusion models~(DMs). In contrast to the under-exploitation of visual characteristics in existing vision tokenizer, we leverage the hierarchical nature of images to encode visual tokens into stratified levels with emergent properties. Through the proposed image stratification that obtains an interlinked token pair, we alleviate the modeling difficulty and lift the generative power of NAR models. Our experiments demonstrate that StraIT significantly improves NAR generation and out-performs existing DMs and AR methods while being order-of-magnitude faster, achieving FID scores of 3.96 at 256$\times$256 resolution on ImageNet without leveraging any guidance in sampling or auxiliary image classifiers. When equipped with classifier free guidance, our method achieves an FID of 3.36 and IS of 259.3. In addition, we illustrate the decoupled modeling process of StraIT generation, showing its compelling properties on applications including domain transfer.
\end{abstract}

\section{Introduction}
\label{sec:intro}

\begin{figure}[tb]
    \centering
    \includegraphics[width=\linewidth]{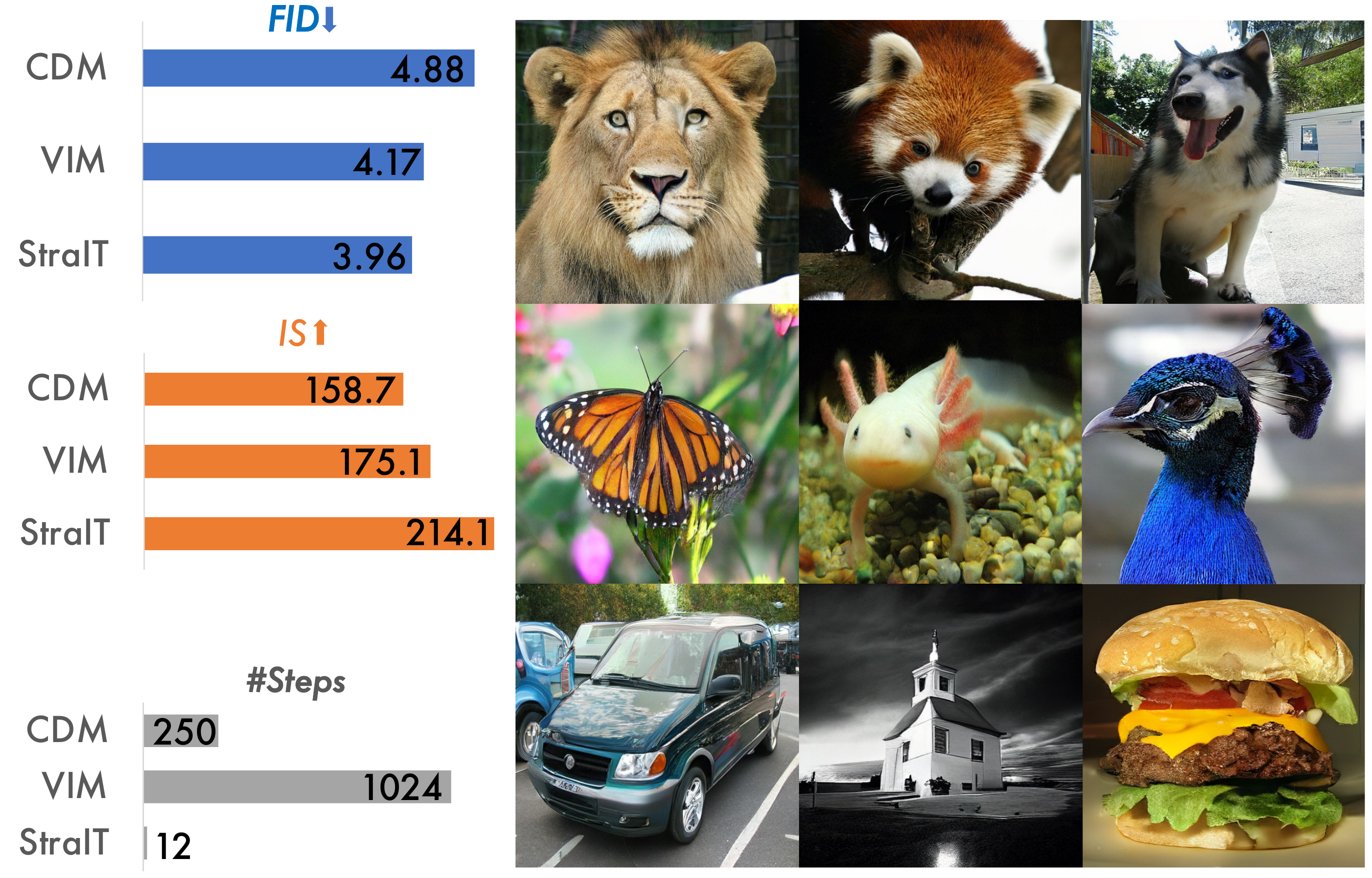}
    \vspace{-0.2in}
    \caption{\small{Quantitative comparison with leading AR and DMs on ImageNet $256^2$ generation: VIM~\cite{ViTVQGAN} and CDM~\cite{CDM}, as well as conditional generated samples from StraIT at the resolution of $512^2$. The methods are compared consistently \textit{without leveraging classifiers, rejection sampling, or classifier free guidance}. More visualizations can be found at Appendix.}}
    \label{fig:teaser}
    \vspace{-0.2in}
\end{figure}

Image generation has recently achieved significant progress, demonstrating prominence in content creation, editing and many other applications. With increasing data and computational resources, leading methods, such as \textit{diffusion models}~\cite{DDPM, ImprovedDDPM, ADM, dalle2, imagen} and \textit{autoregressive transformers}~\cite{VQGAN,parti, dalle, ViTVQGAN}, have largely surpassed prior works based on generative adversarial networks (GANs)~\cite{GAN,sagan,BigGAN,StyleGAN2} in both image quality and diversity. For example, despite being in different model families, diffusion models like ADM~\cite{ADM} and autoregressive models like VIM~\cite{ViTVQGAN} all beat GANs in class-conditional generation. Similarly, DALL-E~\cite{dalle, dalle2}, Parti~\cite{parti} and Imagen~\cite{imagen} have shown unprecedented photorealism compared to GANs for text-to-image synthesis.

Albeit with impressive power, most \textit{autoregressive}~(AR) and \textit{diffusion models}~(DMs) are compute-demanding and have slow sampling speeds, a bottleneck that impedes their accessibility in practical applications. Specifically, DMs commonly require hundreds or thousands of successive evaluation steps in inference, to gradually reduce the noise. AR transformers, on the other hand, need to sequentially decode an image following the raster scan ordering, \ie{~from left to right and line-by-line}. These steps are not parallelizable, resulting in high inference latency. Though several works have explored different strategies~\cite{ddim, lu2022dpm, meng2022distillation, salimans2022progressive} to reduce the sampling steps, they usually sacrifice image quality for faster speed.

\textit{Non-autoregressive transformers}~\cite{maskgit,M6-UFC}, where tokens are decoded in parallel, have contemporarily been explored and demonstrated both promising generation quality and efficiency. For example, MaskGIT~\cite{maskgit} leads to significant inference acceleration while achieving competitive sample quality when trained with the mask-then-predict objective~\cite{BERT} and equipped with iterative decoding~\cite{cmlm}. Despite considerable progress, however, leading \textit{non-autoregressive} models still lag behind state-of-the-art \textit{diffusion}~\cite{CDM} and \textit{autoregressive}~\cite{ViTVQGAN,lee2022draft} counterparts in public benchmarks.

On the other hand, Vector Quantization(VQ)~\cite{vqvae, VQGAN}, which reduces computational complexity through spatial compression, has been widely used in modeling high resolution images. However, a large downsampling rate trades reconstruction quality for efficiency, setting an unavoidable bottleneck for VQ-based generation. LDM~\cite{LDM} recently suggests a relatively mild compression rate for DMs, but it results in much larger latent spatial resolution and longer sequence for modeling in the following stage. The surge in computational costs and memory requirements for handling longer sequences prevent its adoption in NAR transformers. Moreover, recent works in NLP~\cite{guo2019non,gu2017non} also show that long sequence modeling is one of the central challenges of non-autoregressive generation. 

In this paper, we propose \textit{StraIT}, a stratified non-autoregressive model motivated by the actual human painting process. With our tokenizer adaptation, distinctive but interlinked top-level and bottom-level token sequences are obtained from images. Most importantly, this improved token hierarchy intriguingly presents a \textit{short-but-complex} top and \textit{long-but-simple} bottom arrangement reflected in perplexity, relieving the difficulty on modeling longer sequences. With the proposed \textit{Cross-scale Masked Token Modeling} strategy, both the top and bottom-level modules are trained with masked visual token prediction, whereas the second one models the top-to-bottom conditional probability.
 

We make the following three main contributions:

\begin{itemize}
    \item We demonstrate the difficulty on scaling up non-autoregressive model from larger models and longer sequences. To improve NAR generation, we exploit visual characteristics and investigate a suitable tokenization strategy through image stratification.  
    \item With the interlinked token pairs, we propose a stratified modeling framework named StraIT, and empirically demonstrate that, for the first time, our pure NAR method significantly out-performs existing state-of-the-art AR and DMs in on the ImageNet benchmark while achieving 30$\times$ faster inference. 
    \item Furthermore, we conduct extensive ablation studies and provide insights into the generation process of StraIT, demonstrating emergent properties of the decoupled procedure, where the top and bottom transformers own notably different responsibility. We demonstrate that these compelling properties also provide the versatility of StraIT to perform semantic domain transfer, with simple forward passes.
\end{itemize}

\section{Background}
\label{sec:back}
\subsection{Non-autoregressive Image Generation}

It is computationally infeasible to directly model pixel dependencies for high-resolution images. Most of recent \textit{non-autoregressive}~\cite{M6-UFC,maskgit} image generative models adopt the two-stage approach, which consists of a visual tokenization stage and a masked modeling stage. 

\paragraph{Visual Tokenization} In this stage, the goal is to compress the image into discrete and spatially-reduced latent space. The model consists of three major parts: an encoder $E$, a quantizer $Q$ with a learnable codebook $\mathbf{e}$ and a decoder $G$. Given an RGB image $I$ with spatial resolution $(H, W)$, the encoder $E$ first extracts visual features with resolution $(\nicefrac{H}{f}, \nicefrac{W}{f})$, where $f$ is the downsampling ratio. Then the quantizer $Q$ performs a nearest neighbor look-up in the codebook $e$ to quantize latent features into discrete codes. Then the decoder $G$ takes the corresponding features of discrete codes and maps them back to the pixel space to reconstruct the original image. All these three modules are trained together with reconstruction~\cite{vqvae} or adversarial~\cite{VQGAN} objectives. After training, the encoder can extract discrete latent codes for second-stage generative modeling. 


\paragraph{Masked Token Modeling with Transformers}

Similar with the masked language modeling~(MLM) task introduced in BERT~\cite{BERT}, the objective in masked token modeling is also used to predict masked image tokens. Instead of using a fixed masking ratio in language and visual pre-training~\cite{BERT,mae}, the generative transformer needs to generate tokens from scratch and applies a randomly sampled ratio $\gamma (r) \in (0,1]$ in training. Given a sampled binary mask $\mathbf{m}\in \{m_1, ..., m_k\}^K$, the token $y_i$ is replaced with \texttt{[MASK]} if $m_i = 1$, and remains intact when $m_i = 0$. Let $Y_{\overline{\mathbf{M}}}$ denote the masked token sequences. Conditional input $\textbf{c}$, such as class label, is concatenated as a prefix to $Y_{\overline{\mathbf{M}}}$. The objective of training a NAR transformer parameterized by $\theta$ is to minimize the negative log-likelihood of the masked positions:

\begin{equation}
\label{eq:loss}
\mathcal{L}_{\text{mask}}(\theta) = - \mathop{\mathbb{E}} \limits_{\mathbf{Y}  \in \mathcal{D}} \Big[  \sum_{\forall i \in [1,K], m_i=1}\log p(y_i|\textbf{c}, Y_{\overline{\mathbf{M}}}) \Big].
\vspace{-0.2in}
\end{equation}
\begin{figure}[t]
    \centering
    \includegraphics[width=\linewidth]{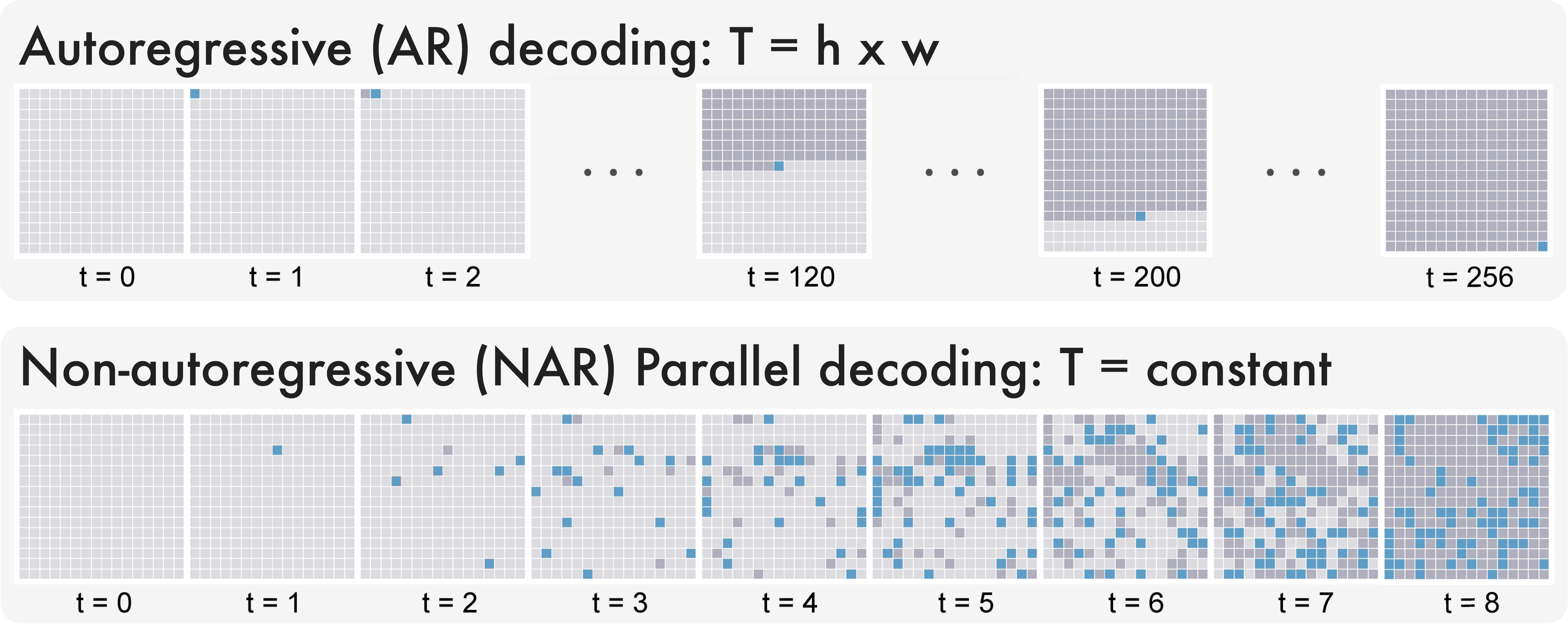}
    \vspace{-0.3in}
    \caption{\textbf{Autoregressive vs Non-autoregressive Decoding.}
    }
    \label{fig:decoding}
    \vspace{-0.2in}
\end{figure}

\begin{figure*}[th]
    \centering
    \vspace{-0.1in}
    \includegraphics[width=0.9\linewidth]{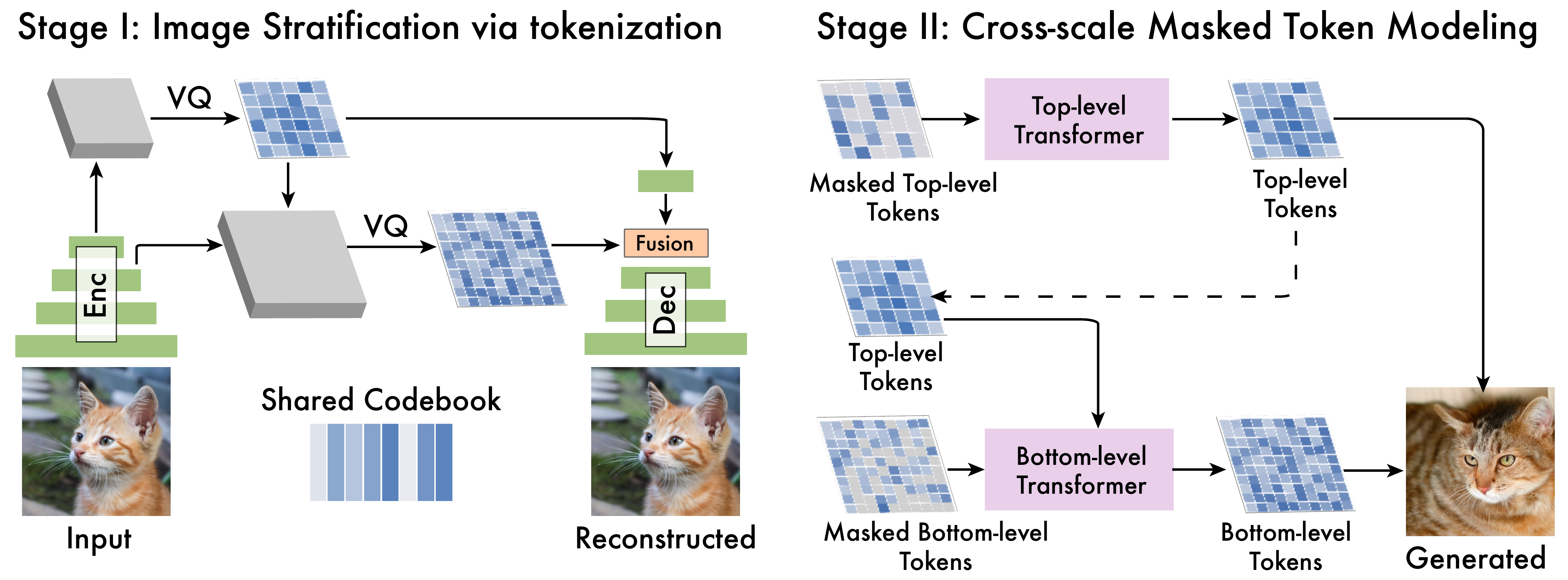}
    \vspace{-0.1in}
    \caption{\textbf{Pipeline Overview. }In the first stage, we obtain stratified visual token sequences using our proposed \textit{VQGAN2-R}. With the \textit{short-but-complex} top and \textit{long-but-simple} bottom token pairs produced by our tokenizer, we design \textit{Cross-scale Masked Token Modeling} to train our top and bottom-level transformers in a decoupled manner. The dotted line denotes a teacher forcing regime in bottom training.}
    \label{fig:pipeline}
    \vspace{-0.1in}
\end{figure*}

\paragraph{Iterative Refinement During Inference}
NAR models can in theory infer all tokens in a single pass, but the performance lags largely behind autogressive generation. Parallel decoding algorithm~\cite{ maskgit,lezama2022improved, lee2022draft, cmlm}, as illustrated in Figure~\ref{fig:decoding}, has been studied in both NLP and vision domains, which offers improved fidelity and diversity while maintains much faster inference than autoregressive counterparts~\cite{ViTVQGAN,VQGAN}. Since our focus is different, we simply adopt the confidence-based strategy in MaskGIT~\cite{maskgit} and propose our stratified iterative decoding in Section~\ref{sec:inference}. 


\subsection{Problems of NAR Image Generation}
\label{sec:improve}
Despite much faster sampling speed, leading NAR methods~\cite{maskgit, lezama2022improved} still under-perform state-of-the-art AR and DMs~\cite{lee2022draft, ADM,ViTVQGAN} in sample quality. Moreover, AR and DMs have achieved much progress by scaling up the architectures~\cite{parti,imagen, CDM} or applying mild downsampling rates to latent spaces~\cite{LDM, ViTVQGAN}. However, we find that directly scaling to larger models (Section~\ref{sec:exp_class}) and longer sequences (Section~\ref{sec:exp_ablation}) for NAR is not sufficient to unravel this performance gap.

\section{Stratified Image Transformer}

In this section, we provide a novel framework called StraIT to improve NAR model for high quality image synthesis.
In Section~\ref{sec:tokenizer}, we study the tokenization step and carefully design \textit{image stratification}, the key technique that enables fine-grained control. In Section~\ref{sec:transformer}, we elaborate our strategy of \textit{decoupled non-autoregressive modeling}.

\subsection{Image Stratification via Tokenization}
\label{sec:tokenizer}

Generative transformers majorly process vision contents in a language-modeling style, where an image~\cite{ding2021cogview, ViTVQGAN,dalle}, video~\cite{ge2022long,hong2022cogvideo}, or other structured inputs~\cite{UnifiedIO,chen2021pix2seq} are processed into a sequence of discrete vocabulary tokens. However, these contents are usually formed with different hierarchies,~\eg{~from subpixels to edges}, which are unfortunately neglected in this style. In this work, we leverage image hierarchy for sequence modeling to achieve better generation results. Namely, we decompose an image into two stratified representations, which inherently reduces the difficulty on modeling long sequences.

Given an image $I \in \mathbb{R}^{H \times W \times 3}$, our tokenizer outputs two spatial collections of codebook entries, $\mathbf{Y}_{top}$ and $\mathbf{Y}_{bottom}$, where $\mathbf{Y}_{top} \in \mathbb{R}^{\frac{H}{16} \times \frac{W}{16}}$ and $\mathbf{Y}_{bottom} \in \mathbb{R}^{\frac{H}{8} \times \frac{W}{8}}$. They are then flattened to 1D, resulting in a short top sequence $Y_{t} = [y_i^t]_{i=1}^N$ and a long bottom sequence $Y_{b} = [y_i^b]_{i=1}^{4N}$.

\paragraph{Stratified Tokenization} As depicted in Figure~\ref{fig:pipeline}, we utilize a two-level token hierarchy. For  $256 \times 256$ images, the encoder $E$ first transforms and downsamples the inputs by factors of 8 and 16, obtaining feature representations of $32 \times 32$ and $16 \times 16$ respectively. The $16 \times 16$ latent map is firstly quantized to our top-level codes. The latent conditional layer, which consists of several residual blocks, upscales the quantized top-level map and then stack it with the $32 \times 32$ features. After the final bottom-level quantization, we obtain $Y^t$ and $Y^b$, two stratified token sequences with different lengths. 

For the encoder structure, we follow the design choice in VQVAE-2~\cite{vqvae2}. While for the decoder applied with different fusion strategies, we observe distinctive representations. 

To differentiate, we adopt two variants: \textit{VQGAN2-C}, which follows the strategy in VQVAE-2~\cite{vqvae2}, fusing top and bottom levels together by concatenation; and \textit{VQGAN2-R}, which processes the bottom level as residuals with~\textit{stratified residual fusion},~\ie{~firstly upsample the top-level features, and then add the bottom-level features onto it}.



\paragraph{Training Objectives.} Following VQGAN~\cite{VQGAN}, we apply the perceptual loss~\cite{johnson2016perceptual} and adversarial loss~\cite{VQGAN} in seeking perceptual quality. For simplicity, we use the shared codebook to quantize top and bottom features, where the commitment loss~\cite{vqvae} is applied to both layers. Thus, the total vector quantization training loss is, 

\vspace{-0.2in}
\begin{equation}
\label{eq:lossvq}
\mathcal{L}_{\text{VQ}}(E, G, \mathbf{e}) = \mathcal{L}_{\text{Adv}} + \mathcal{L}_{\text{Perc}} + \mathcal{L}_{\text{Commit}}(Y_b, Y_t).
\vspace{-0.1in}
\end{equation}

\begin{table}[htb]
\begin{center}
\resizebox{\columnwidth}{!}{%
\begin{tabular}{l|cc|cc}
\toprule
\multirow{2}{*}{Model} & \multirow{2}{*}{$\#$Tokens} & \multirow{2}{*}{FID/PSNR} & \multicolumn{2}{c}{\textbf{Perplexity~(PPL)}} \\ \cline{4-5} 
                       &                         &                                  & \textbf{Top}           & \textbf{Bottom}         \\ \hline
VQGAN~($f$=16)~\cite{VQGAN}                  & $16^2$                      & 2.04/19.9                & \multicolumn{2}{c}{5893 $\pm$ 2.1}       \\
VQGAN~($f$=8)~\cite{VQGAN}                  & $32^2$                      & 0.81/24.3              & \multicolumn{2}{c}{6387 $\pm$ 2.8}       \\ \hline
\textit{VQGAN2-C}              & $16^2+32^2$                    & 0.65/24.9                  & 405 $\pm$ 1.3        & 6428  $\pm$ 4.1        \\
\textit{VQGAN2-R}               & $16^2+32^2$                     & 0.67/24.8                 & 5632  $\pm$ 2.0       & 1644   $\pm$ 1.9         \\ \bottomrule
\end{tabular}}
\end{center}
\vspace{-0.2in}
\caption{\label{table:vqgan}Comparison between \textit{VQGAN2-C} and \textit{VQGAN2-R}, as well as our re-implemented VQGAN~($f$=8,16) with a consistent recipe. All four models have a codebook size of 8192. }
\vspace{-0.1in}
\end{table}

\paragraph{Fusion Strategy and Emergent Properties.} We provide quantitative evaluation of our stratified tokenizer in Table~\ref{table:vqgan}, and qualitative comparison in Appendix~\ref{appendix:tokenization}. Compared to single-scale tokenization, \textit{VQGAN2-C} and \textit{VQGAN2-R} perform slightly better by introducing extra tokens. To understand the distinction between the tokens learned by them, we report the per-batch perplexity~(PPL) during training. Serving as a one of the most common metrics for evaluating language models, VQ perplexity measures the codebook utilization~\cite{takida2022sq} and reflects the complexity of sequences: 

\begin{itemize}
    \item \textit{VQGAN2-C}: the model exhibits a `greedy' property, where the bottom level has a high PPL and top level has an extremely low PPL, showing it is rarely exploited. 
    \item \textit{VQGAN2-R}: conversely, the bottom level works as an residual to the top level, making the top level has a much higher PPL than bottom. 
\end{itemize}

While previous works such as VQVAE-2~\cite{vqvae2} adopted hierarchical codes for more powerful priors over the latent codes, our focus is to obtain a suitable conditional distribution for non-autoregressive modeling on longer sequences. In other words, we want to relieve the burdens on modeling longer sequences non-autoregressively. Accordingly, we adopt \textit{VQGAN2-R} as it provides interlinked \textit{short-but-complex} top and \textit{long-but-simple} bottom token representations. We also provide detailed comparisons in Section~\ref{sec:exp_ablation}.

\subsection{Cross-scale Masked Token Modeling}
\label{sec:transformer}
With stratified tokenizer $E$ and de-tokenizer $G$ available, we can now represent an image with two dependent sequences. In contrast to previous sequence modeling framework, we propose to learn two decoupled transformers by \textit{Cross-scale Masked Token Modeling}~(CMTM). In the following, we illustrate the process, training objectives, and inference techniques. 

\paragraph{Decoupled Modeling}

Let $Y^t = [y_i^t]_{i=1}^N$ denote the obtained top-level tokens, where $N$ is the length of the flatten matrix, and $Y^b = [y_i^b]_{i=1}^{4N}$ represents the respective bottom-level tokens. Instead of separate or joint modeling, we choose to model them in a decoupled manner: 
\begin{itemize}
    \item \textit{Top-level transformer}: serves as a likelihood model that tries to generate top-level tokens purely from scratch.
    \item \textit{Bottom-level transformer}: models the conditional likelihood, and learns to predict the corresponding bottom-level tokens given top-level inputs. 
\end{itemize}

\paragraph{Training Objectives} We adopt \textit{Cross-scale Masked Token Modeling} to train the top and bottom-level model parameterized by $\theta^t$ and $\theta^b$. For the top-level transformer, the task is to predict masked tokens directly. While for the bottom-level transformer, strong conditional guidance is provided by top-level tokens, which makes the modeling process easier. Let $\mathbf{m}^t$ and $\mathbf{m}^b$ denote the independently sampled masks. The top-level and bottom-level masked token modeling losses are: 
\begin{equation}
\label{eq:loss_top}
\mathcal{L}_{\text{mask}}^{\text{top}}(\theta^t) =-\mathop{\mathbb{E}} \limits_{\mathbf{Y^t}  \in \mathcal{D}} \Big[  \sum_{\substack{m_i^t=1, \\ \forall i \in [1,N]}} \log p(y_i^t| \textbf{c}, Y_{\overline{\mathbf{M}}}^t) \Big],
\vspace{-0.1in}
\end{equation}
\begin{equation}
\label{eq:loss_bottom}
\hspace{-0.1in}
\mathcal{L}_{\text{mask}}^{\text{bot}}(\theta^b) = -\mathop{\mathbb{E}} \limits_{\mathbf{Y^b}  \in \mathcal{D}} \Big[\sum_{\substack{m_i^b=1, \\ \forall i \in [1,4N]}}\log p(y_i^b|\textbf{c},Y^t, Y_{\overline{\mathbf{M}}}^b) \Big].
\vspace{-0.05in}
\end{equation}
In practice, these two objectives enable separate training of the top and bottom-level transformers, reducing much memory cost. Such simple yet effective stratified modeling doesn't require conditional augmentations~\cite{CDM}, which we elaborate more in Appendix~\ref{appendix:transformer}.

\subsection{Inference with StraIT}
\label{sec:inference}

\paragraph{Stratified Iterative Decoding}
We follow the iterative parallel decoding in CMLM~\cite{cmlm} and MaskGIT~\cite{maskgit} to generate images. In contrast to previous non-autoregressive methods, we generate two stratified sequences in a top-down manner. Specifically, the top-level transformer predicts all tokens starting from a blank canvas $\overline{M}$ where all tokens are masked out. Each refinement step fills the canvas with a number of tokens according to their predicted probability. The completely predicted top-level sequence then guides the bottom-level transformer to perform its conditional iterative decoding on $\overline{N}$ following a similar procedure. 

Our decoding process is illustrated as follows:

\begin{algorithmic}[1]
{\small 
\Require{$\overline{M},\overline{N}\,{=}\,\varnothing$ , $T^\text{top}$, $T^\text{bottom}$, $\gamma$} 
\For {$t \gets 1$ to $T^\text{top}$}
\State $n\,=\,\gamma(t, T^\text{top})$,  $\hat{y}_{i}^t\,{\sim}\,P_{\theta^t}(y_{i}^t|\widehat{\mathcal{Y}^t}_{\overline{M}}, \textbf{c})$, $\forall i\,{\in}\,{M}$
\State $\overline{M} \gets \overline{M}\cup \{\arg\mathrm{topk}_{i\,{\in}\,M}\big(P_{\theta^t}(y_{i}^t|\widehat{\mathcal{Y}^t}_{\overline{M}}, \textbf{c}), k\,{=}\,n\big)\}$
\EndFor
\For {$t \gets 1$ to $T^\text{bottom}$}
\State $n\,=\,\gamma(t, T^\text{bottom})$,  $\hat{y}_{i}^b\,{\sim}\,P_{\theta^b}(y_{i}^b|\widehat{\mathcal{Y}^b}_{\overline{N}}, \overline{M}, \textbf{c})$, $\forall i\,{\in}\,{N}$
\State $\overline{N} \gets \overline{N}\cup \{\arg\mathrm{topk}_{i\,{\in}\,N}\big(P_{\theta^b}(y_{i}^b|\widehat{\mathcal{Y}^b}_{\overline{N}}, \overline{M},\textbf{c}), k\,{=}\,n\big)\}$
\EndFor
}
\end{algorithmic}
where $T^\text{top}$, $T^\text{bottom}$ denote the steps of top and bottom-level decoding. For $\gamma$, we adopt a cosine function from~\cite{maskgit}. We study the allocation of decoding steps in Section~\ref{sec:exp_ablation}.

\section{Experiments}
\label{sec:exp}
In this section, we evaluate the performance of StraIT on image generation, in terms of quality-diversity, efficiency, and adaptability. In Section~\ref{sec:exp_class}, we provide both quantitative and qualitative evaluations on the standard class-conditional image generation on ImageNet~\cite{ImageNet}. In Section~\ref{sec:exp_ablation}, we conduct ablation studies to understand our stratified modeling process, as well as showing its advantages over different variants. Then in Section~\ref{sec:exp_app}, we analyze the intriguing property of our system, and show its compelling applications.

\subsection{Experimental Setup}
For all experiments, we train our tokenizer \textit{VQGAN2-R} with a single codebook of 8192 tokens using cropped $256\times256$ images from ImageNet~\cite{ImageNet}. The images are always downsampled by factors of 16 and 8, respectively producing top and bottom-level tokens. The codebook is also used to train our model on $512\times512$. 

For \textit{Cross-scale Masked Token Modeling}, we leverage two different transformer architectures, including the top-level transformer that consists of self-attention blocks and the bottom-level transformer that adopts the cross-attention blocks. Sharing embedding dimensions of 768, the models used in this work have the following configuration:

\begin{table*}[t!]
\begin{center}
\resizebox{0.8\textwidth}{!}{%
\begin{tabular}{llcccccccc}
\toprule
Family                          & Method         & \# Params   & \# Steps & FID $\downarrow$  & IS  $\uparrow$ & Precision $\uparrow$  &  Recall $\uparrow$ &  \\
\midrule
\multirow{4}{*}{Autoregressive}  & VQVAE-2~\cite{vqvae2}         & 13.5B     & 5120   & 31.11 & 45  & 0.36   &  0.57  \\
                & VQGAN~\cite{VQGAN}         & 1.3B     & 256   & 15.78 & 78.3  & -   &  -  \\
                & RQ-Transformer~\cite{rqtransformer} & 3.8B     & 256  & 7.55  & 137  &  -  &  -  \\
                                & VIT-VQ+VIM~\cite{ViTVQGAN} & 1.7B     & 1024 & 4.17  & 175.1   &     -   & -  \\
\midrule
\multirow{6}{*}{Diffusion Models}      &  Improved DDPM~\cite{ImprovedDDPM} & 280M     & 250   & 12.26 & - & 0.70 &  0.62 \\
                                & ADM~\cite{ADM}            & 554M     & 250   & 10.94 & 101.0  & 0.69  & 0.63 \\
                        & VQ-Diffusion~\cite{gu2022vector}   & 518M     & 100   & 11.89 & -   &  -   &  -   \\
                                & LDM~\cite{LDM}            & 400M     & 250   & 10.56 & 103.49 & 0.71 & 0.62 \\
                                & CDM~\cite{CDM}            & $\sim$1B & 250  & 4.88  & 158.71 &  -   &  -  \\
\midrule
\multirow{3}{*}{Non-autoregressive}  & MaskGIT~\cite{maskgit}         & 227M     & 8    & 6.18  & 182.1  & 0.80 & 0.51  \\
     & MaskGIT$^\dagger$  & 1.3B     & 12    & 5.84  & 180.3 &  0.73 &  0.54 \\
     & MaskGIT$^\dagger$  & 1.3B     & 36    & 5.71  & 185.9 &  0.73 &  0.56 \\
                 & \bf StraIT        & 863M     & 12$^\ast$    & \bf 3.96  & \bf 214.1  & 0.74  &  0.62 \\
\bottomrule
\end{tabular}}
\end{center}
\vspace{-0.2in}
\caption{\label{table:exp_class}Quantitative comparison with state-of-the-art generative models on ImageNet $256\times256$ \textbf{without leveraging any guidance or external classifiers for training and inference}. For VQ-based methods, `\# Params' includes the parameters of VQ. `\# Steps' denotes the number of forward runs to generate a sample. $^\dagger$ denotes the re-implementations with the same setup with ours. $^\ast$ our method adopts an (18+6)-step allocation, which is faster than a 12-step inference of our whole model. Details are provided in Sec.~\ref{sec:exp_ablation} and Table~\ref{table:exp_step}.}
\end{table*}

\begin{itemize}
    \vspace{-0.05in}
    \item Top: 48 layers, an intermediate size of 5120, and 32 attention heads, leading to 499M parameters.
    \vspace{-0.2cm}
    \item Bottom: 16 layers, an intermediate size of 4096, and 24 attention heads, leading to 292M parameters.
    \vspace{-0.05in}
\end{itemize}

We train baselines and two models proposed using AdamW~\cite{loshchilov2017decoupled} with a base learning rate of 1e-4, adopting a 5000-step linear warmup and a cosine decaying schedule afterward. Label smoothing~\cite{szegedy2016rethinking} and dropout~\cite{srivastava2014dropout} are also employed following~\cite{maskgit}. Each model is trained for 200 epochs with a batch size of 256 on TPU chips. To allow fair comparisons and investigate influence from larger models, we adopt our consistent recipes to train MaskGIT~\cite{maskgit} with 1.3B parameters, which receives marginal improvement shown in Table~\ref{table:exp_class} and indicates the inefficiency to naively scaling model sizes in existing NAR paradigms.

\subsection{Main Results on Image Synthesis}
\label{sec:exp_class}

Table~\ref{table:exp_class} and Table~\ref{table:exp_guidance} summarize the main results of StraIT for $256 \times 256$ class-conditional generation on ImageNet~\cite{ImageNet}. Since improved inference strategies have been explored for different generative models, it's difficult to compare them in a unified way. Therefore, we conduct comparisons without any guidance in training and inference in Table~\ref{table:exp_class} and then incorporate classifier-free guidance~\cite{cfg} to StraIT, reporting the results in Table~\ref{table:exp_guidance}.

\begin{table}[htb]
\begin{center}
\resizebox{\columnwidth}{!}{%
\begin{tabular}{lcccc}
\toprule
Method         & Params & Steps  & FID $\downarrow$  & IS  $\uparrow$   \\
\midrule
\bf w. Learnable guidance \\
Improved VQ-Diffusion~\cite{tang2022improved} & 510M & 100  & 4.83 & -   \\
Token Critic~\cite{lezama2022improved} & 391M & 18 $\times$ 2 & 4.69 & 174.5   \\
DPC~\cite{DPC} & 391M & 180  & 4.45 & 244.8   \\
\midrule
\bf w. Classifier-free guidance\\
\textit{f}-DM~\cite{gu2022f} & 302M & 250  & 6.8  & -    \\
Draft-and-Revise~\cite{lee2022draft} & 1.4B & 68$\times$(1+4)$^\dagger$  & 3.41  & 224.6    \\
\bf StraIT            & 863M  & (18+6)$^\dagger$ & \bf 3.36 & \bf 259.3    \\

\bottomrule
\end{tabular}}
\end{center}
\vspace{-0.2in}
\caption{\label{table:exp_guidance}Comparison on methods with improved inference strategies, where StraIT adopts a classifier-free guidance~\cite{cfg} scale of 0.2. $\dagger$ \cite{lee2022draft} adopts a light depth transformer for the 4 steps in (1+4).}
\end{table}

\paragraph{Quantitative Evaluation} From Table~\ref{table:exp_class}, we show that, without any special sampling methods, our method significantly out-performs previous state-of-the-arts in both Fr\'{e}chet Inception Distance (FID)~\cite{FID} and Inception Score~(IS). For the first time, we demonstrate that non-autoregressive model out-performs advanced autoregressive~\cite{ViTVQGAN,rqtransformer} and diffusion~\cite{CDM} models with much fewer steps. In addition, our method maintains a trade-off between precious and recall, which suggests a better coverage (Recall) compared to MaskGIT~\cite{maskgit}, and improved sample quality~(Precision) to diffusion models.

While our focus is not to introduce improved sampling methods, we incorporate the commonly used classifier free guidance~\cite{cfg} by DMs to our method. Specifically, we adopt a probability of randomly dropping conditioning in training as $0.1$, and a guidance scale of 0.2 during inference. As demonstrated in Table~\ref{table:exp_guidance}, our method achieves the best FID and IS on ImageNet generation on record when not using classifiers or rejection sampling, further showing its strength.

\paragraph{Higher Resolution in Token-level} To demonstrate the versatility of non-autoregressive model, we follow recent cascade diffusion models~\cite{ADM,LDM,CDM} to generate higher resolution~\ie{ $512\times512$} in a purely non-autoregressive manner. For simplicity, we utilize our same tokenizer trained on $256\times256$. Using the same architecture of bottom-level transformer, we train another upsampling model to generate the stratified tokens representing $512\times512$ images conditioned on the lower resolution tokens obtained by bilinear downsampled $256\times256$ images. Due to the memory cost, we adopt a simple three-layer model denoted by $U$. During inference, the generated sequences from StraIT is fed into $U$ for iterative decoding to tokens for $512\times512$ outputs. 

We provide results on ImageNet $512\times512$ conditional generation in Table~\ref{table:exp_512}. Our simple three-layer model performs significantly better than ADM-$U$~\cite{ADM} that adopts a more intensive upsampling strategy. Notably, it also out-performs ADM-$U$-G in terms of IS, which adopts pre-trained classifier guidance. Note that this comparison does not intend to push performance on $512\times512$, rather suggesting the prominence of pure NAR generation.

\begin{table}[htb]
\begin{center}
\resizebox{0.8\columnwidth}{!}{%
\begin{tabular}{lcccc}
\toprule
Method         & $T_{\text{base}}$ & $T_{\text{upsample}}$  & FID $\downarrow$  & IS  $\uparrow$   \\
\midrule
Token Critic~\cite{lezama2022improved} & 36 & - & 6.80 & 182.1 \\
\midrule
ADM~\cite{ADM}  & 250 & - & 23.24  & 58.06    \\
ADM-$U$  & 250 & 250 & 9.96  & 121.78    \\
ADM-$U$-G & 250 & 250 & 3.85  & 221.72    \\ \midrule
StraIT-$U$ & 12 & 12  & \bf 3.82  & \bf 253.6    \\
\bottomrule
\end{tabular}}
\end{center}
\vspace{-0.2in}
\caption{\label{table:exp_512}Results of $512\times512$ image generation on ImageNet. $T_{\text{base}}$ and $T_{\text{upsample}}$ denote the number of step to perform lower-scale generation and upsampling, respectively.}
\vspace{-0.15in}
\end{table}

\paragraph{User Preference Study} To showcase the capabilities of NAR formalism, we follow CDM~\cite{CDM} to compare with other methods directly, \ie{~avoid using external image classifiers to boost sample quality}. However, recent works like StyleGAN-XL~\cite{stylegan-xl} that adopted classifier guidance achieve promising quantitative results. To allow more systematic and reliable comparisons, we conduct user preference studies using Amazon Mechanical Turk on verifying the quality and diversity of generated samples from four leading methods on ImageNet $512\times512$ generation: StyleGAN-XL~\cite{stylegan-xl} and ADM-$U$-G~\cite{ADM} that adopt classifier guidance, as well as MaskGIT~\cite{maskgit} and Token Critic~\cite{lezama2022improved}. 

For \textit{quality} evaluation which selects the more realistic one, each grader is presented with two randomly sampled images of a same class side-by-side, one using StraIT, the other using one of the competing methods. For \textit{diversity} evaluation, two batches of twelve generated images from same classes are provided in the same way, where the graders choose the more diverse-looking one. For both quality and diversity, each pairwise comparison is rated by 15 graders.


\begin{figure}[thb]
    \centering
    \vspace{-0.14in}
    \includegraphics[width=1\linewidth]{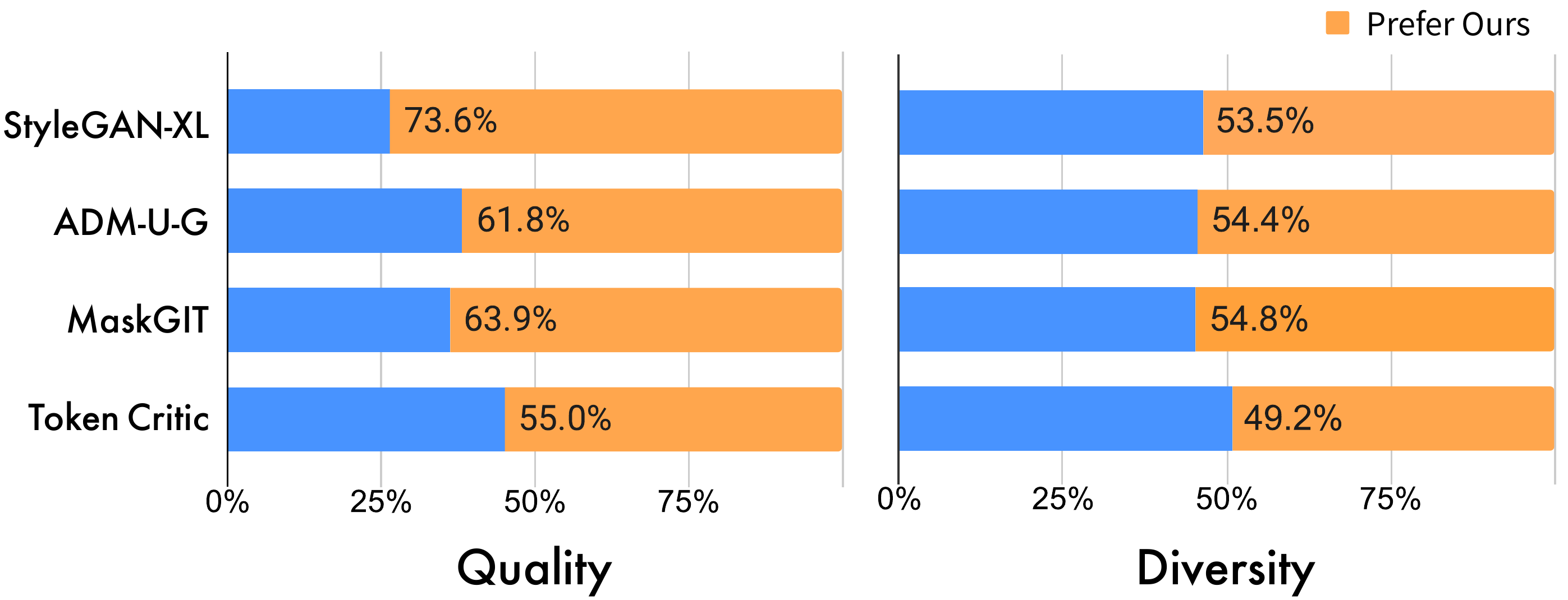}
    \vspace{-0.1in}
    \caption{\textbf{User preference study} on quality and diversity. The number denotes the portion of times when StraIT is preferred over competing methods.}
    \label{fig:exp_userstudy}
    \vspace{-0.1in}

\end{figure}

As shown in Table~\ref{table:exp_512}, our method is more preferred than all other competing methods in terms of \textit{quality}. Comparing to these existing best-performing models on ImageNet generation ranging from GANs, transformers, and diffusion models, our method shows a clear advantage despite the simple structure. For \textit{diversity}, it appears that existing methods are capable of generating diverse samples, which are relatively harder for graders to distinguish. However, our method still shows favoured \textit{diversity} over state-of-the-art GAN and DMs that exploit pre-trained classifier: StyleGAN-XL~\cite{stylegan-xl} and ADM-$U$-G~\cite{ADM}, as well as leading generative transformers~\cite{maskgit,lezama2022improved}.

\subsection{Ablation Studies}
\label{sec:exp_ablation}
To better understand our stratified process, we conduct ablation studies and provide detailed illustration. For consistency, we don't use classifier-free guidance in this section.

\paragraph{Scaling up NAR on Sequence Lengths} As investigated in LDM~\cite{LDM}, aggressive spatial compression,  \ie{~$f$=16} in VQGAN~\cite{VQGAN}, significantly eliminates high frequency in images. Therefore, applying relatively mild downsampling rates would notably improve quality in diffusion model. By replacing the original $f$ = 16 VQGAN with $f$ = 8 in MaskGIT~\cite{maskgit} and with increased training and inference costs, we show in Table~\ref{table:sequence} that the final performance is conversely decreased. We conjecture that the conditional independence assumption of NAR models becomes more problematic when dealing with longer sequences. This result also serves as the motivation of our stratified modeling.

\begin{table}[htb]
\begin{center}
\resizebox{0.8\columnwidth}{!}{%
    
\begin{tabular}{ccccc}
\toprule
\# Tokens & FLOPs & Training Costs & FID $\downarrow$ & IS $\uparrow$ \\ \midrule
$16 \times 16$  & 48.2    & $1\times$    &  6.05 & 203 \\
$32 \times 32$ &  216.5   & $2.6\times$          & 6.61 &  177.7  \\ \bottomrule
\end{tabular}}
\end{center}
\vspace{-0.2in}
\caption{\label{table:sequence} {Results of MaskGIT~\cite{maskgit} model trained with different sequence lengths. Training costs are measured on same TPUs.}}
\vspace{-0.2in}
\end{table}

\paragraph{\textit{VQGAN2-R} over \textit{VQGAN2-C}} While they share identical architectures apart from the fusion strategy, the proposed \textit{VQGAN2-R} behaves distinctively with \textit{VQGAN2-C}. As discussed in Section~\ref{sec:tokenizer} and Table~\ref{table:vqgan}, different fusion strategy leads to distinctive token representations. Most importantly, we want to reduce modeling complexity on longer sequences, therefore adopting the \textit{long-but-simple} visual tokens produced by \textit{VQGAN2-R}. To tell the difference, we provide the results by replacing our tokenizer with \textit{VQGAN2-C} and keep other architectures consistent. In Table~\ref{table:exp_ablation}, we show that \textit{VQGAN2-C} leads to much worse FID scores, suggesting that such simple modification deteriorates the transformer significantly. This gap also indicates the importance of proper tokenizers~\cite{VQGAN} for generative vision transformer, which has been rarely studied.

\paragraph{\textit{Stratified} over \textit{Cascade}} Another option to improve longer sequences modeling is to use a cascade pipeline that generates sequences progressively, similarly as the upsampling strategy~\cite{CDM,ADM} in DMs. To consistently compare in $256\times256$ resolution, we utilize two tokenizers of VQGAN($f$=8, 16) from Table~\ref{table:vqgan} to construct a cascade variant. Using a similar pipeline in our CMTM, we replace the top and bottom-level tokens with $16\times16$ ones from VQGAN($f$=16) and $32\times32$ ones from VQGAN($f$=8). As this top-level code can generate images using its own decoder, we report the results on ImageNet $256\times256$ in Table~\ref{table:exp_ablation} from both the whole system and top-level transformer. We find that the modified cascade pipeline helps improve the results of $f$=16 baseline marginally, expressing the validity of adopting guidance for longer sequences from shorter ones. In contrast to this coarse-to-fine design, our stratified tokenizer extracts hierarchical and interlinked visual token sequences, which significantly benefit \textit{Cross-scale Masked Token Modeling}.

\begin{table}[htb]
\begin{center}
\resizebox{0.95\columnwidth}{!}{%
\begin{tabular}{cccccc}
\toprule
\multirow{2}{*}{Type}       & \multirow{2}{*}{Tokenizer} & \multicolumn{2}{c}{Transformers} & \multirow{2}{*}{FID $\downarrow$} & \multirow{2}{*}{$\Delta$} \\ \cline{3-4}
                            &                            & Top        & Bottom        &                      &                   \\ \hline
\multirow{2}{*}{Stratified} & \textit{VQGAN2-C}                  &      $\checkmark$      &       $\checkmark$        &    6.11              &         -          \\
                            & \bf \textit{VQGAN2-R}                   &    $\checkmark$        &     $\checkmark$          & \bf    3.96          &     -2.15              \\ \hline
\multirow{2}{*}{Cascade}    & VQGAN~($f$=16)                  &    $\checkmark$        &               &           6.05           &                   \\
                            & VQGAN~($f$=8)                       &  $\checkmark$           &       $\checkmark$         &   \bf     5.71              &     -0.34              \\ \bottomrule
\end{tabular}}
\end{center}
\vspace{-0.2in}
\caption{\label{table:exp_ablation}Results of $256\times256$ image generation on ImageNet from different tokenizers and pipelines. The results from Cascade $f$=16 are obtained by decoding predicted tokens from Top-level.}
\vspace{-0.2in}
\end{table}

\paragraph{Decoding Steps} Unlike previous generative transformers~\cite{gpt,maskgit,BERT}, our paradigm results in two decoupled sequences. Following the iterative decoding strategy~\cite{cmlm} and cosine mask scheduling~\cite{maskgit}, we study the effect from different allocations on decoding steps in Table~\ref{table:exp_step}. Interestingly, spending more decoding steps on the top level benefits both FID and IS. Considering the teacher forcing training regime on the bottom level, it makes sense to `pay attention to the condition'. It's note-worthy that the top model, despite with more parameters, operates on short $16\times16$ sequences and requires less computation than the bottom-level architecture which operates on $32\times32$ sequences~\ie{~top: 133.5G FLOPS, bottom: 311.3G FLOPS}. As it does not reflect actual inference speed, we estimate the wall-clock time over 50000 generation samples. Given better results and notable speedup, we adopt $18+6$ decoding steps for our experiments.

\begin{table}[htb]
\begin{center}
\resizebox{0.7\columnwidth}{!}{%
\begin{tabular}{ccccc}
\toprule
$T^{\text{top}}$  & $T^{\text{bottom}}$  & FID $\downarrow$ & IS $\uparrow$ & Speedup \\
\midrule
3  & 21 &   5.4  &  193  &    $0.6\times$    \\
6  & 18 &   5.6  & 192   &      $0.7\times$   \\
9  & 15 &  4.5   &  201.3  &   $0.9\times$      \\ \midrule
12 & 12 &  4.21   &  202.7  & $1\times$      \\ \midrule
15 & 9  &   4.05  &  214  &   $1.2\times$       \\
18 & 6  & \bf 3.97   & \bf 214.1    &   $1.5\times$       \\
21 & 3  &  4.0   &  209  &  $1.6\times$   \\
\bottomrule
\end{tabular}}
\end{center}
\vspace{-0.2in}
\caption{\label{table:exp_step}Comparisons on decoding steps. $T^{\text{top}}$ and $T^{\text{bottom}}$ represent the top-level and bottom-level decoding steps. Speedup is estimated by generating 50000 images on TPUv3.}
\vspace{-0.2in}
\end{table}

\begin{figure*}[t]
    \centering
    \includegraphics[width=0.9\linewidth]{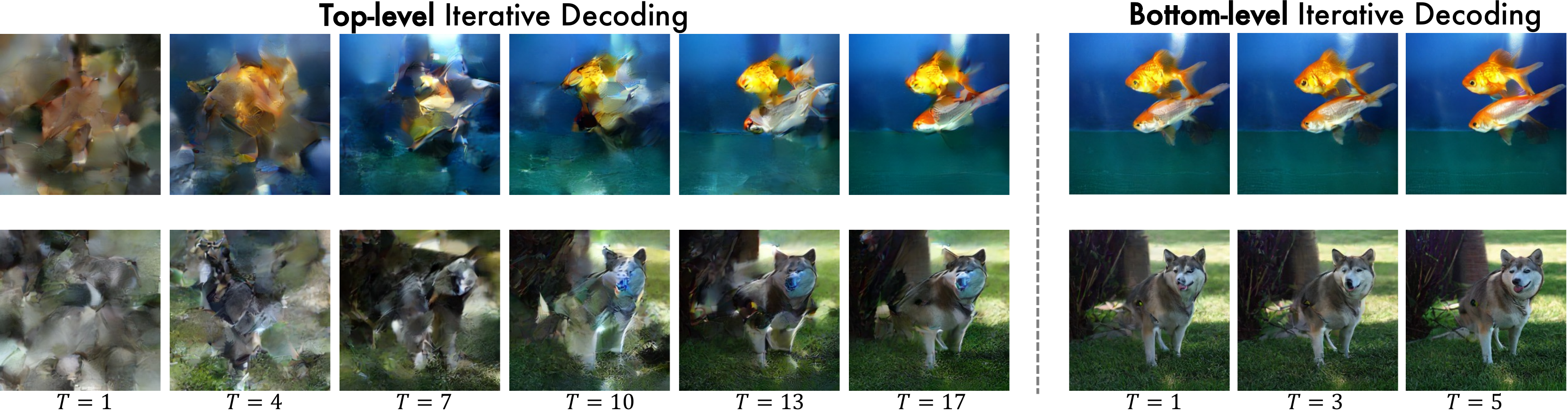}
    \vspace{-0.1in}
    \caption{\textbf{Intermediate outputs at different steps of our stratified Iterative Decoding.} \small{ $T$ on the left and right denotes the current step of top and bottom-level decoding. For visualizing the top-level iterative decoding process while the bottom-level code is not available, we provide a fixed randomly-sampled bottom-level tokens to the decoder.}}
    \label{fig:step}
    \vspace{-0.05in}
\end{figure*}

\subsection{Intriguing Properties and Applications}
\label{sec:exp_app}

In this section, we first illustrate the stratified decoding process of StraIT, showing an intriguing decoupled property. Then we show the versatility of StraIT on doamin transfer, without any architecture changes or fine-tuning.

\paragraph{Stratified Modeling Process} To better understand our stratified modeling process, we further provide visualizations of the inference steps. As shown in Figure~\ref{fig:step}, the top-level transformer gradually infills basic colors, coarse boundaries, and general layout of an image, while the bottom-level model refines detailed textures to these regions and produces realistic images. This intriguing phenomenon empirically explains the stratified information encoded in our top-level and bottom-level visual tokens. Moreover, the decoupled generation process is analogous to human painting, where an image draft is firstly constructed by rough brushstrokes, and then the bottom-level transformer, serving like an expert editor, gradually embroiders the entire image.

Non-autoregressive models trained with \textit{Masked Token Modeling} have a nature of learning to infill. Therefore, they can be seamlessly applied to multiple image editing tasks by handling them as constraints to the input mask. Previous works on NAR model~\cite{maskgit} have demonstrated its versatility of image inpainting and extrapolation. Recently, Phenaki~\cite{villegas2022phenaki} also adopts NAR for variable length video generation. Our work serves as a stronger NAR framework and also share similar advantages. Instead of echoing these benefits, we show emerging applications from our framework.

\begin{figure}[htb]
    \centering
    \includegraphics[width=\linewidth]{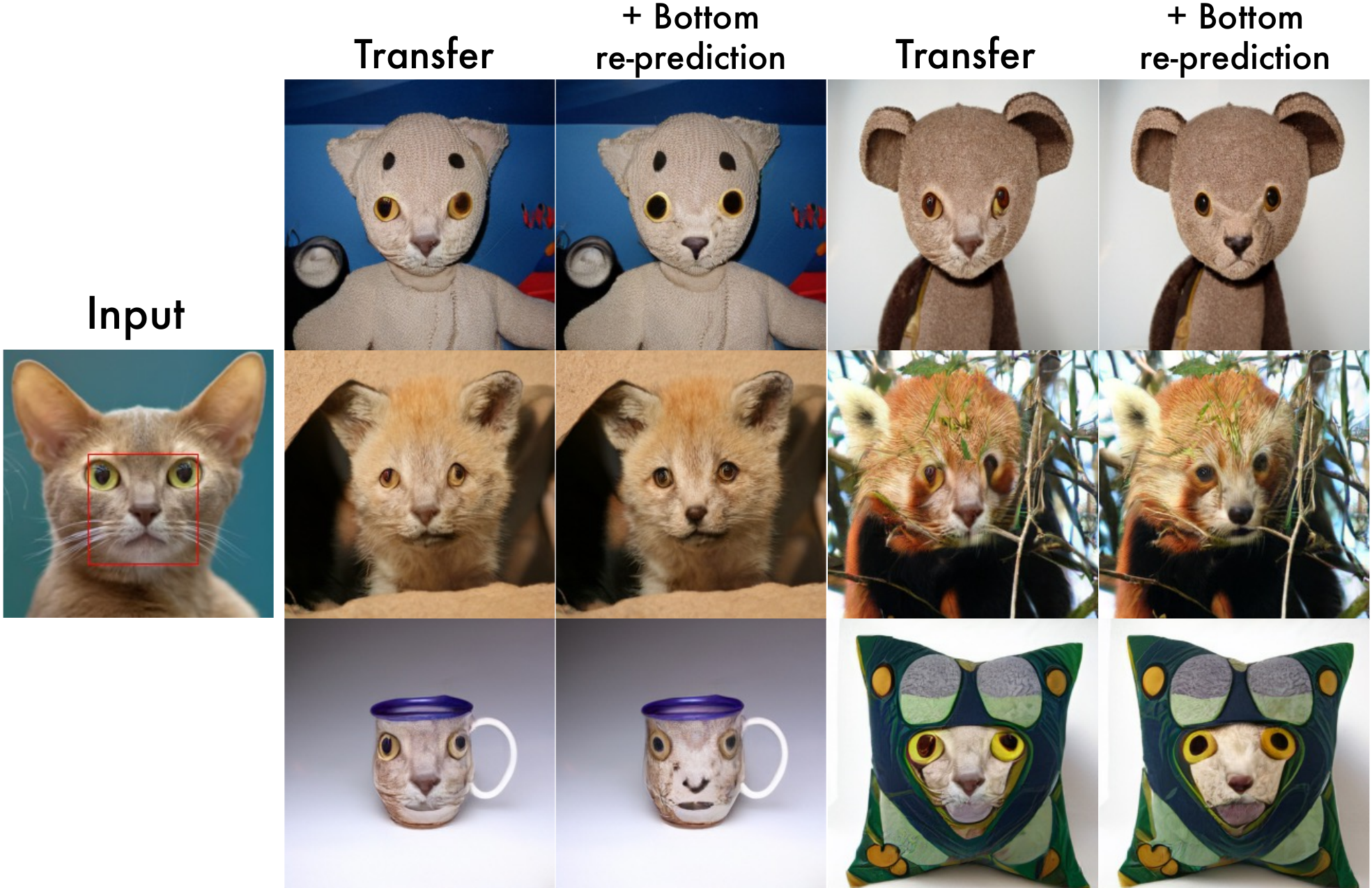}
    \vspace{-0.3in}
    \caption{\textbf{Transferring a cat to different semantic domains}. Targeted domains include: \texttt{[Teddy]}$\times2$, \texttt{[Kit Fox]}, \texttt{[Red Panda]}, \texttt{[Coffee Mug]}, \texttt{[Pillow]}. \textbf{Zoom in for details.}}
    \label{fig:transfer}
    \vspace{-0.1in}
\end{figure}

\paragraph{Semantic Domain Transfer} One compelling benefit of StraIT is to perform domain transfer. Since StraIT enables decoupled modeling which provides better semantic abstraction, we can now perform domain transfer easily by masking unwanted tokens. As shown in Figure~\ref{fig:transfer}, our method successfully transfers the chosen geometry and skeleton to diverse domains, even with large semantic variance. 

\paragraph{Bottom-level Re-prediction} As validated in Figure~\ref{fig:step}, the bottom-level transformer is designed to perform detailed refinement given the top-level visual tokens, akin to the original texture. To allow transfer adaptability, one simple strategy is to mask all bottom-level tokens and leverage the transformer to perform re-prediction from the transferred top-level tokens. With no extra computation incurred, such simple strategy benefits domain transfer clearly and naturally, from the comparison in Figure~\ref{fig:transfer}. In supplement to Figure~\ref{fig:step}, the results further elaborate our stratified modeling: the top level performs semantic understanding and generate coarse layout; while the bottom level edits and re-touches on the visual details. 

These intriguing properties of StraIT open up many possibility of applying NAR to generation and editing, especially considering its fast decoding and simple formalism. 

\section{Related Works}

\paragraph{Visual Tokenization} By converting an image into a sequence of
discrete codes, Vector Quantization~\cite{vqvae, dalle} has been adopted to obtain visual tokens. Recent works such as VQGAN~\cite{VQGAN} and ViT-VQGAN~\cite{ViTVQGAN} try to maintain fine details with improved techniques, while RQ-VAE~\cite{rqtransformer, lee2022draft}, on the other hand, represents the image as a stacked map of discrete codes. Prior works have also investigated hierarchical VQ for learning powerful priors~\cite{vqvae2,transformer_token}, representing multi-level textures for human synthesis~\cite{jiang2022text2human}, or to attain high factors of compression~\cite{williams2020hierarchical}.

\paragraph{Non-autoregressive Generation} While recent generative models, including AR and DMs, show improved results on both images~\cite{imagen,parti,dalle2} and videos~\cite{imagenvideo}, another family of non-autoregressive models~\cite{gu2017non, maskgit,M6-UFC, villegas2022phenaki} with accelerated inference have also been explored. Different from recent works~\cite{DPC,lezama2022improved} that investigated better sampling strategies, our work targets at proposing a new NAR paradigm that handles longer sequences. Recently, \cite{lee2022draft} incorporated AR with NAR generation, showing another direction of hybird NAR modeling.

\section{Conclusion}
In this work, we propose StraIT, a stratified non-autoregressive generative paradigm that out-performs existing autoregressive and diffusion model on class-conditional image generation. The proposed strategies of image stratification and \textit{Cross-scale Masked Token Modeling} allow predicting longer visual tokens sequences that include more fine-grained details. The promising results, order-of-magnitude faster inference, and versatility to applications make StraIT an attractive choice for generative modeling. Our studies indicate that the non-autoregressive family is a promising direction for future research.

\section*{Acknowledgements}
\noindent We would like to thank Jos\'e Lezama for discussions and generous suggestions, as well as Ruben Villegas for the helpful reviews and constructive feedback on the draft.

{\small
\bibliographystyle{ieee_fullname}
\bibliography{egbib}
}

\newpage

\appendix

\newcommand{\suppsection}[1]{\section{{\fontsize{12}{13}\selectfont #1}}}

\noindent The content of the appendix is organized as follows:

\begin{itemize}
    \item Qualitative results from StraIT in Appendix.~\ref{appendix:visual}
    \item Discussion on Stratified Tokenization in Appendix.~\ref{appendix:tokenization}
    \item Discussion on Decoupled Transformer in Appendix.~\ref{appendix:transformer}
    \item More experimental details in Appendix.~\ref{appendix:details}
    \item Limitations and future works in Appendix.~\ref{appendix:limitation}
\end{itemize}

\suppsection{Qualitative Results from StraIT}
\label{appendix:visual}

In this section, we provide more qualitative results to show the effectiveness of StraIT.

\begin{figure*}[htb]
    \centering
    \caption{\small{Selected $512\times512$ generated samples from StraIT.}}
    \vspace{-0.15in}
    \includegraphics[width=0.84\linewidth]{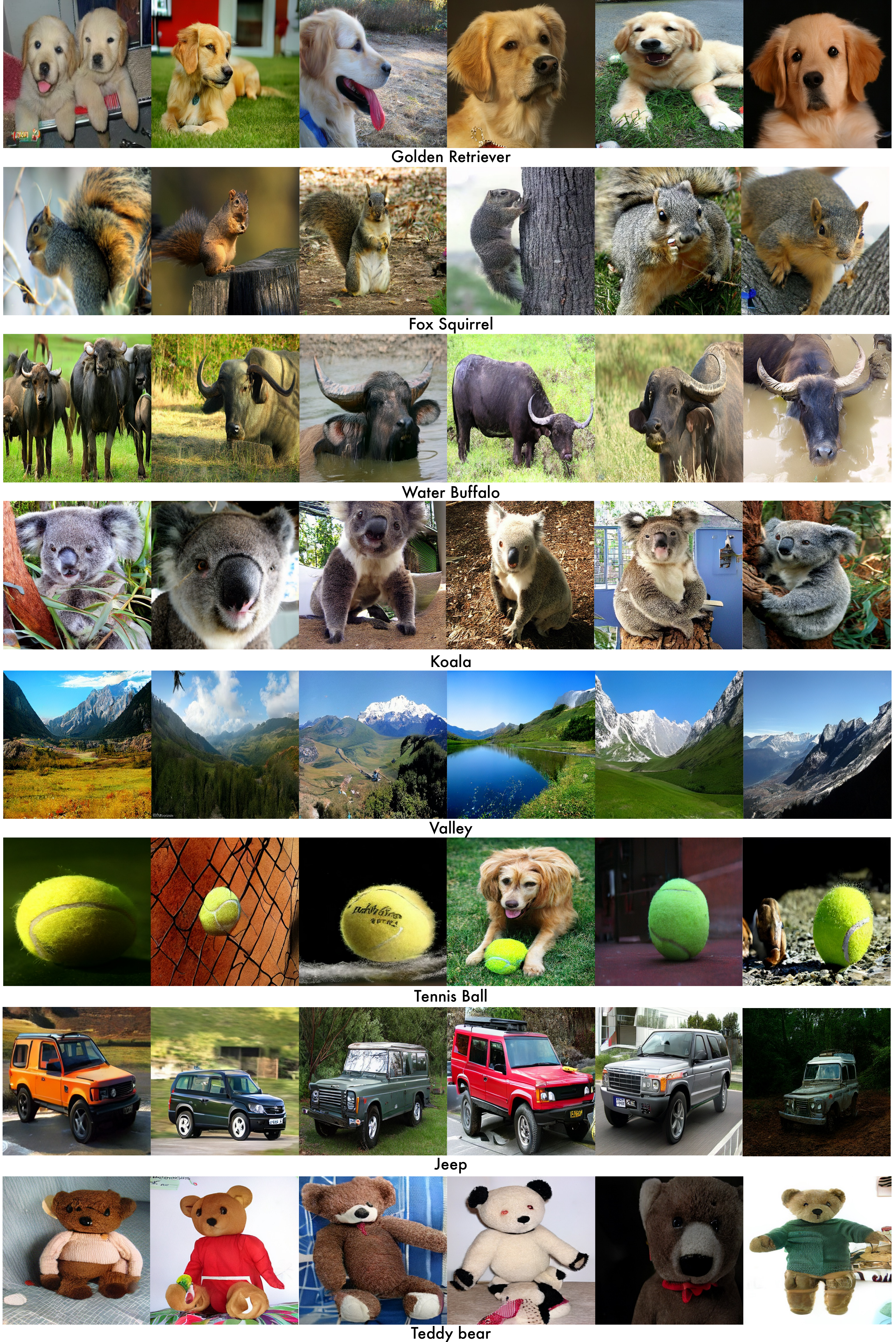}
    \label{fig:supp_class_1}
    \vspace{-0.1in}
\end{figure*}

\begin{figure*}[htb]
    \centering
    \caption{\small{Selected $512\times512$ generated samples from StraIT. }}
    \vspace{-0.15in}
    \includegraphics[width=0.84\linewidth]{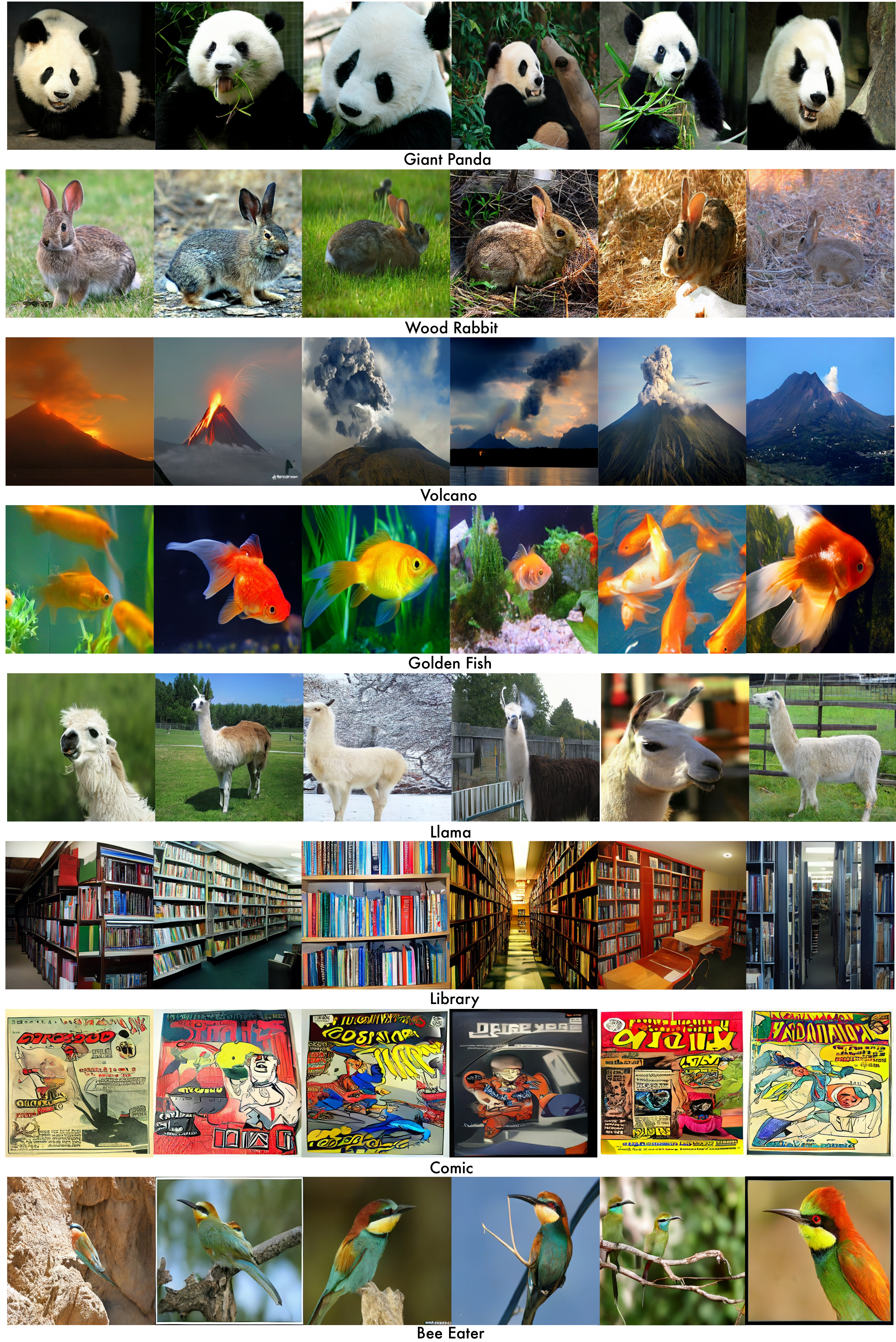}
    \label{fig:supp_class_2}
    \vspace{-0.1in}
\end{figure*}

\paragraph{Image Generation} We provide $512\times512$ image generation results of different classes from ImageNet in Figure~\ref{fig:supp_class_1} and~\ref{fig:supp_class_2}. In order to demonstrate the sample quality and diversity, we provide multiple samples given one class. With much smaller number of inference steps, StraIT provides diverse samples with high visual quality.

\begin{figure*}[htb]
    \centering
    \vspace{-0.15in}
    \includegraphics[width=\linewidth]{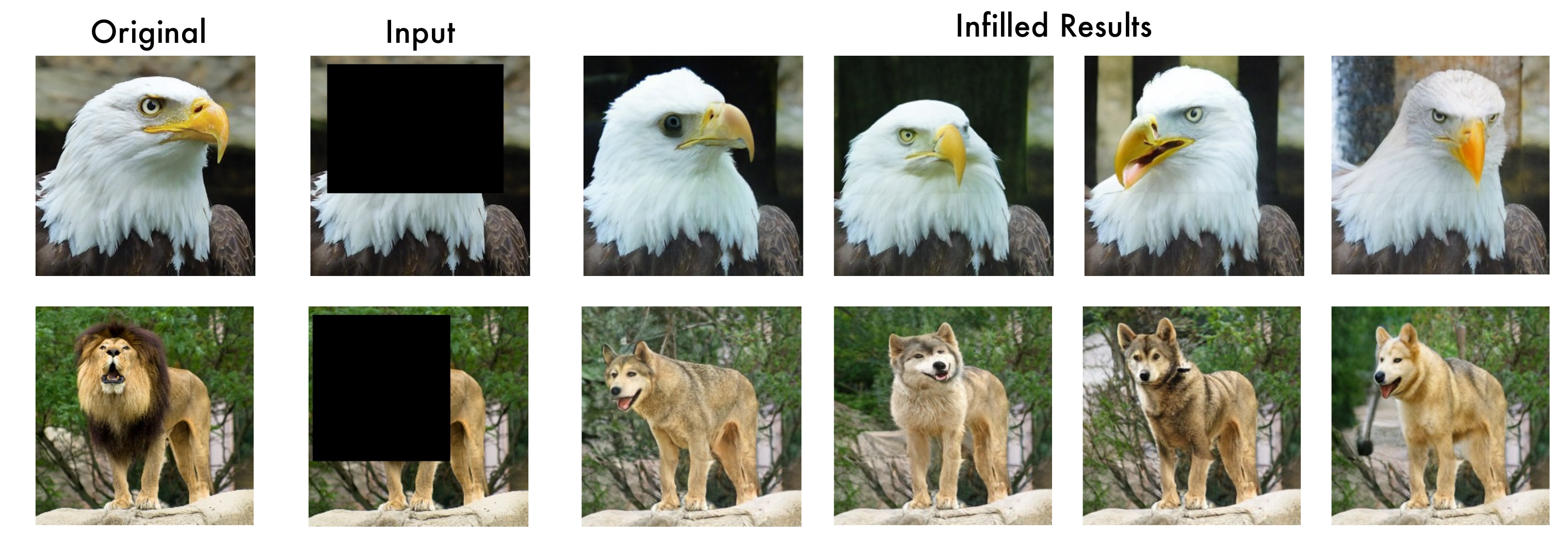}
    \caption{\small{Flexible image infilling result. The second image in each row denotes the input image to our model. The first row represents inpainting results with the same given label. The second row shows the infilled results from different context~\ie{~turning lion into husky}.}}
    \label{fig:inpainting}
    \vspace{-0.1in}
\end{figure*}

\paragraph{Flexible Image Editing} Since NAR models trained with masked token modeling have the nature of infilling missing contents, StraIT is capably of editing image flexibly in simple feedforward passes by tokenizing unmasked tokens. We provide diverse image editing results in Figure~\ref{fig:inpainting}, including infilling missing regions and replacing with other context.

\suppsection{Discussion on Stratified Tokenization}
\label{appendix:tokenization}

Image stratification, which represents an image with two distinctive sequences of visual tokens, is an important design in our work. We investigate modified tokenizations in order to obtain stratified visual codes~\ie{~the top-level codes decide coarse layout and color distribution with thick strokes, while the bottom level stores refined texture}. Such stratified nature further enables~\textit{Cross-scale Masked Token Modeling} and leads to the decoupled generation process of StraIT.

\paragraph{Perplexity} To compare variants of tokenizers, we adopt the perplexity~(PPL) to measure the complexity of visual tokens. While the perplexity metric has been widely used in measuring language models, it denotes codebook utilizations of tokenizers as well as the complexity of token representations in vision. The perplexity is computed as: 

\begin{equation}
\label{eq:perplexity}
\text{Perplexity}(p) = 2^{-\sum_x {p(x)\log_2{p(x)}}} = \prod_{x} {p(x)^{-p(x)}}
\end{equation}

Note that we report the per-batch PPL during training, which converges after 700k iterations. The stabilized per-batch PPL indicates the ordinary complexity of visual tokens. 

\begin{figure}[htb]
    \centering
    \vspace{-0.1in}
    \includegraphics[width=\linewidth]{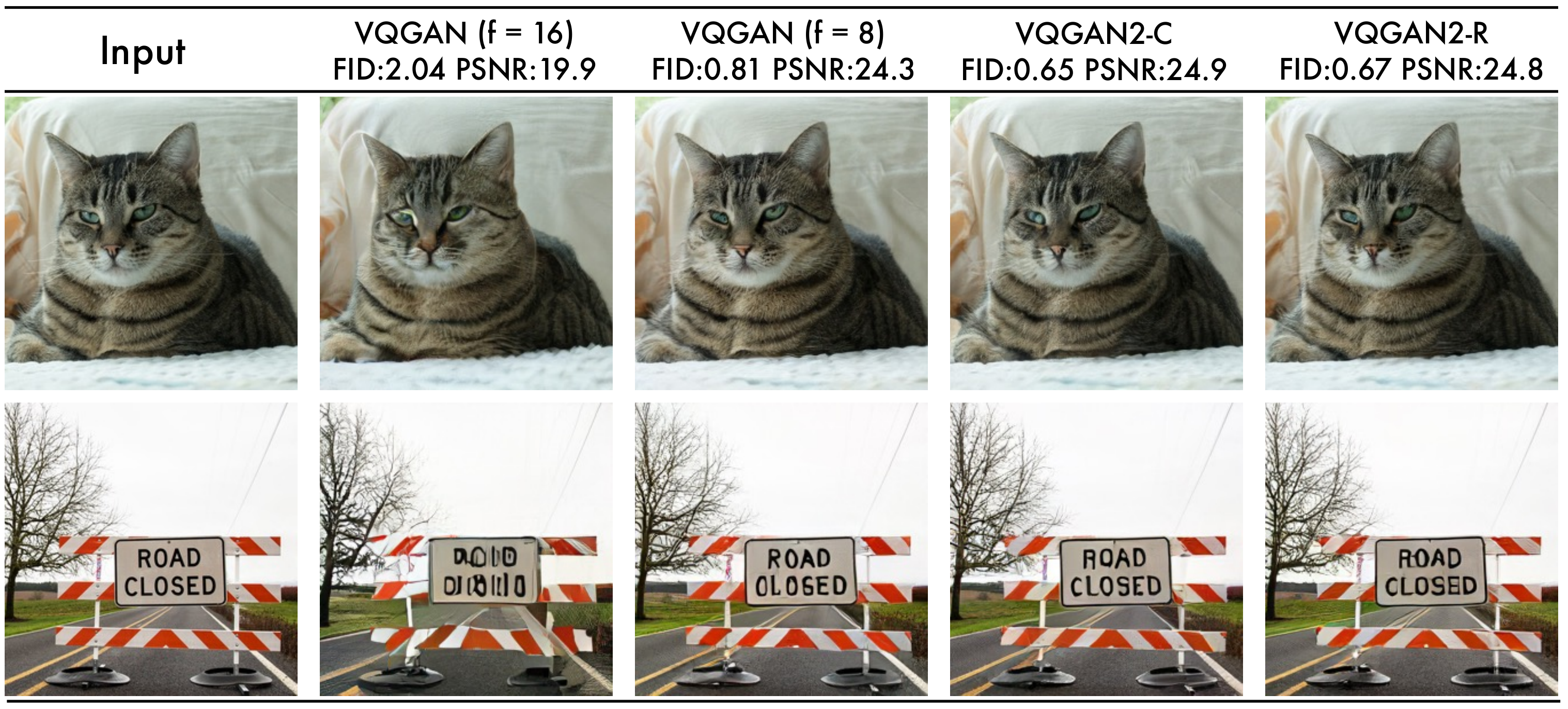}
    \caption{Qualitative comparisons among different tokenizers.}
    \label{fig:vqgan_compare}
    \vspace{-0.1in}
\end{figure}

\paragraph{Comparison between \textit{VQGAN2-R/C}} As mentioned, we propose image stratification to provide suitable token sequences for non-autoregressive modeling. The experiment results in Table~\textcolor{blue}{5} show that, with longer sequences, NAR transformer actually performs worse than using short ones. While naively increasing sequence length shows inferior performance, we choose to leverage the hierarchical nature in images by representing them into interlinked token pairs. With the guidance from top-level codes, we expect our bottom-level transformer to perform better on modeling longer sequences. However, when training the tokenizer, different fusion strategies in decoder lead to distinctive token representations, as have been quantified in Table~\textcolor{blue}{1}.

\begin{itemize}
    \item \textit{VQGAN2-C}, on the one hand, concatenates the bottom level~($f$=8) and the upscaled top level~($f$=16) together, acting greedily to exploit bottom-level code which provides more information with less spatial compression. Similarly as VQVAE2~\cite{vqvae2}, this bottom-level emphasis leads to poor exploitation in top-level codes, which has also been observed in recent work~\cite{jiang2022text2human}.
    \item \textit{VQGAN2-R}, in contrast, decodes directly from top level and treats the bottom level as residual by adding it back to the second level of de-tokenizer. Such simple-yet-effective \textit{stratified residual fusion} strategy leads to stratified visual tokens, as the model processes top-level visual codes as stem and relies less on the bottom level.
\end{itemize}

The differences in perplexity and the visualization of our generation process have reflected the difference between top and bottom-level visual codes. To provide more context, we also conduct qualitative comparison in Figure~\ref{fig:vqgan_compare}. It's noteworthy that our target is not to build VQ variants with stronger capability or better reconstructions. As shown, both VQGAN~($f$=8) and \textit{VQGAN2} perform better than $f$=16, and \textit{VQGAN2-R} performs slightly worse than \textit{VQGAN2-C}. However, \textit{VQGAN2-R} provides more suitable visual tokens for decoupled non-autoregressive modeling, where the \textit{short-but-complex} top and \textit{long-but-simple} bottom arrangement gives much better generation performance.

\paragraph{Separate Codebooks} Since our experiments have shown the difference between top and bottom tokens, a possible option is to tokenize each level with a separate codebook that represents dedicated features. Considering the perplexity of~\textit{VQGAN2-R}, we replace the shared $\|\mathcal{Z}\| = 8192$ codebook with two separate codebooks of $\|\mathcal{Z}_{\text{top}}\| = 8192$ and $\|\mathcal{Z}_{\text{bottom}}\| = 2048$. Following the consistent experimental settings, we report the performance with separate codebooks in Table~\ref{table:sep_codebook}. The results show that, though with increased number of tokens, using separate codebook only improves the quantitative results slightly. For simplicity, we adopt the shared codebook for top and bottom level in our experiments. 

\begin{table}
\begin{center}

\resizebox{0.8\columnwidth}{!}{%
\begin{tabular}{c|cc|cc}
\toprule
\multirow{2}{*}{Method} & \multicolumn{2}{c|}{Codebook Size $\mathcal{Z}$} & \multirow{2}{*}{FID~$\downarrow$} & \multirow{2}{*}{IS~$\uparrow$} \\ \cline{2-3}
                        & $\|\mathcal{Z}_{\text{top}}\|$               & $\|\mathcal{Z}_{\text{bottom}}\|$              &                      &                     \\ \hline
Shared                  & \multicolumn{2}{c|}{8192}              &            3.96          &           214.1          \\
Separate                &         8192         &        2048         &          3.89            &           216.3          \\ \bottomrule
\end{tabular}}
\end{center}
 \renewcommand\thetable{8}
 \vspace{-0.1in} 
 \caption{Results on ImageNet $256\times256$ generation with shared and separate codebook designs, without classifier free guidance.}
 \label{table:sep_codebook}
\end{table}

\suppsection{Discussion on Decoupled Transformer}
\label{appendix:transformer}

As have been discussed, the transformers in StraIT model the stratified visual tokens in a decoupled manner using~\textit{Cross-scale Masked Token Modeling}. We've also illustrated the respective roles of top and bottom-level transformer in visualizations. Beyond the adaptations we've proposed in the main paper, we provide additional investigation on this modeling process.

\paragraph{Conditional Augmentation} When training the bottom-level transformer, we adopt a teacher forcing training regime~\ie{~the ground truth top-level code $Y^t$ is used directly as the training condition}. However, during inference, the top-level condition is generated from our top-level model, where domain gaps might exist. Such gaps in conditions have been proven to significantly harm the generation results in GANs~\cite{zhang2017stackgan} and cascaded diffusion model~\cite{CDM,imagen}. To reduce this gap, we further study conditional augmentation in training our decoupled architectures.

Different from the practice in cascaded diffusion~\cite{CDM}, where the low resolution images are degraded with random permutations to reduce potential exposure bias, the generated top-level sequences in StraIT are semantic-aware. To make the bottom-level model more robust towards varying top-level code during inference, we experiment to randomly mask a subset of the ground truth top-level code, and then exploit the top-level model $\theta^t$ to predict the contents on partially-masked regions. Let $\widetilde{Y^t}$ denote the augmented conditions, the objective function becomes:

\begin{equation}
\label{eq:loss_bottom_aug}
\hspace{-0.1in}
\mathcal{L}_{\text{mask}}^{\text{bot}}(\theta^b) = -\mathop{\mathbb{E}} \limits_{\mathbf{Y^b}  \in \mathcal{D}} \Big[\sum_{\substack{m_i^b=1, \\ \forall i \in [1,4N]}}\log p(y_i^b|\textbf{c},\widetilde{Y^t}, Y_{\overline{\mathbf{M}}}^b) \Big].
\vspace{-0.05in}
\end{equation}

We study the effect from different strengths~\ie{~masking ratios} of conditional augmentations in Table~\ref{table:cond_aug}. In contrast to the conditional augmentations in diffusion models~\cite{CDM,imagen} that play a key role, sample quality and diversity of StraIT doesn't receive benefits from them. As NAR and DMs appear to behave differently, we choose to not include such augmentations in our framework. The straight-forward teacher forcing regime shows state-of-the-art results. 

\begin{table}
\begin{center}
\resizebox{0.8\columnwidth}{!}{%
\begin{tabular}{c|c|cc}
\toprule
Method    &  Masking ratio  &  FID~$\downarrow$  &  IS~$\uparrow$ \\
\hline
 Original   & -  & \bf 3.96  & \bf 214.1  \\
\midrule
\multirow{3}{*}{Augmented}    &  10\% & 4.03  & 212.8  \\
    & 30\%  & 4.05  & 208.7  \\
    & 50\%  &  3.99 &  213.5 \\
\bottomrule
\end{tabular}}
\end{center}
 \renewcommand\thetable{9}
 \vspace{-0.1in} 
 \caption{Results from different masking ratios adopted in conditional augmentation.}
 \label{table:cond_aug}
\end{table}

\paragraph{Mask Scheduling Functions} One key design in NAR generation is iterative decoding. Note that we do not focus on finding better inference algorithms in this work, thus following the choice in MaskGIT~\cite{maskgit} to use a cosine function in determining mask scheduling~\ie{~the fraction of tokens decoded each iteration}. To better understand this process, we provide ablation studies on different functions in Table~\ref{table:function}. As shown, concave functions generally produce better results than convex ones, suggesting the necessity on using proper decoding techniques.

\begin{table}
\begin{center}
\resizebox{0.7\columnwidth}{!}{%
\begin{tabular}{c|c|cc}
\toprule
Function    &  $\gamma$  &  FID~$\downarrow$  &  IS~$\uparrow$ \\
\hline
 \multirow{2}{*}{Convex}   & Logarithmic & 11.32  & 167.3  \\
    & Square Root  & 7.89  & 187.5  \\ \midrule
   - & Linear  &  5.13 &  200.3 \\ \midrule
 \multirow{4}{*}{Concave}    & Cosine  & \bf 3.96 &  \bf 214.1 \\
  & Square  &  4.13 &  209.5 \\
  & Cubic  &  4.97 &  203.1 \\
  & Exponential  &  4.09 &  212.7 \\
\bottomrule
\end{tabular}}
\end{center}
 \renewcommand\thetable{10}
 \vspace{-0.1in} 
 \caption{Ablation studies on different mask scheduling functions. Each model uses a (18+6)-step. We report the best FID and IS.}
 \label{table:function}
\end{table}

\suppsection{Additional Experimental Details}
\label{appendix:details}
In this section, we further provide more implementation details about architectures, hyper-parameters, and user preference study.

\paragraph{Implementation Details} We provide the detailed model architectures and hyper-parameters of \textit{VQGAN2-R} in Table~\ref{table:param_vq}. This recipe is consistently leveraged in training other VQGAN variants in our experiments. For our top and bottom-level transformers, we share the same training settings, as shown in Table~\ref{table:param_trans}. Both the tokenizer and transformers are trained on 32 TPU chips. For all model variants, we report their best FID and IS by sweeping the sampling temperatures~\cite{maskgit}. 

\begin{table}[htb]
\begin{center}\begin{tabular}{c|c} 
\toprule
Hyper-parameters     &   \textit{VQGAN2-R}   \\ \hline
training epochs &   200  \\  \hline
batch Size &   256   \\
optimizer &   SGD   \\
learning rate &   1e-4   \\
lr schedule &   constant   \\
gradient penalty  &  R1 reg~\cite{mescheder2018training}  \\
penalty cost &    10.0  \\  
commitment cost & 0.25 \\
GAN loss weight   &    0.1 \\
perceptual loss weight & 0.1  \\  \hline
codebook size      &   8192 \\
embedding dim~\cite{ViTVQGAN}     &   32 \\
activation      &   Swish~\cite{swish} \\
normalization      &   Group Norm~\cite{GN} \\
\#channels     &     128      \\
\#res blocks     &     2      \\
channel multi.  &  $[1, 1, 2, 2, 4]$ \\  \hline
discriminator        &   StyleGAN d~\cite{StyleGAN2}   \\
norm in d        &   Group Norm~\cite{GN}  \\
\#channel multi. of d      &    $[1]$       \\
blur resample      &    $\checkmark$     \\
\bottomrule
\end{tabular}
\end{center}
\renewcommand\thetable{11}
\caption{\label{table:param_vq}Model architectures and hyper-parameters of \textit{VQGAN2-R}. The training process is performed on 32 TPUv4 chips.}
\end{table}

\begin{table}[htb]
\begin{center}\begin{tabular}{c|c} 
\toprule
Hyper-parameters     &   Transformers   \\ \hline
training epochs &   200  \\  \hline
batch Size &   256   \\
optimizer &   AdamW~\cite{loshchilov2017decoupled}   \\
learning rate &   1e-4   \\
weight decay &  0.045  \\
momentum &  $\beta_1, \beta_2 = 0.9, 0.96$  \\
gradient clip &  3.0  \\
label smoothing~\cite{szegedy2016rethinking} &   0.1 \\
warmup steps &   5000   \\
uncond cutoff~\cite{cfg} &     0.1  \\
\bottomrule
\end{tabular}
\end{center}
\renewcommand\thetable{12}
\caption{\label{table:param_trans}Hyper-parameters of the decoupled transformers in StraIT. The top and bottom-level model share the same training recipes.}
\end{table}

\begin{figure*}[htb]
    \centering
    \vspace{-0.1in}
    \includegraphics[width=0.9\linewidth]{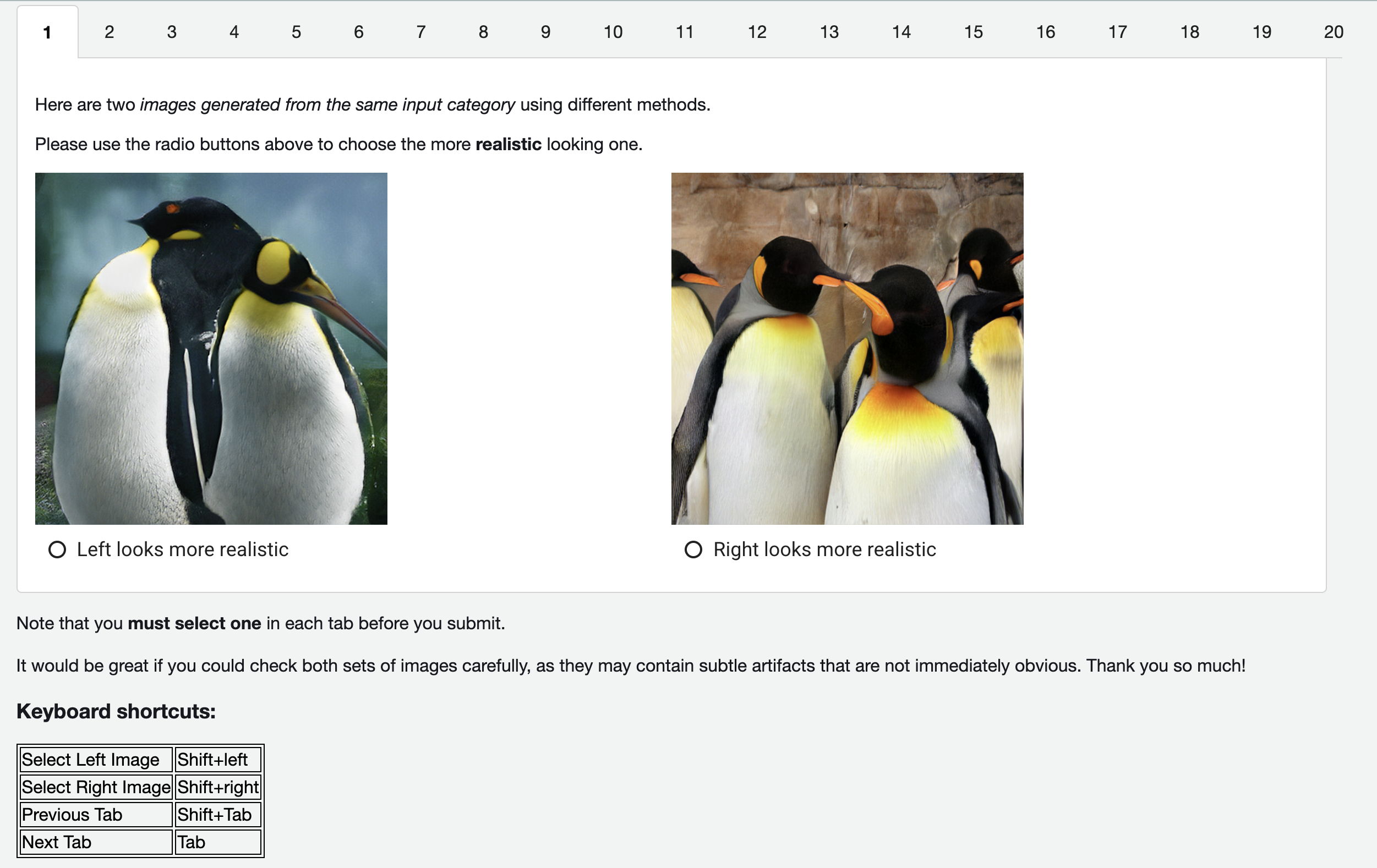}
    \caption{User Interface of quality evaluation.}
    \label{fig:quality_eval}
    \vspace{-0.1in}
\end{figure*}

\begin{figure*}[htb]
    \centering
    \includegraphics[width=0.9\linewidth]{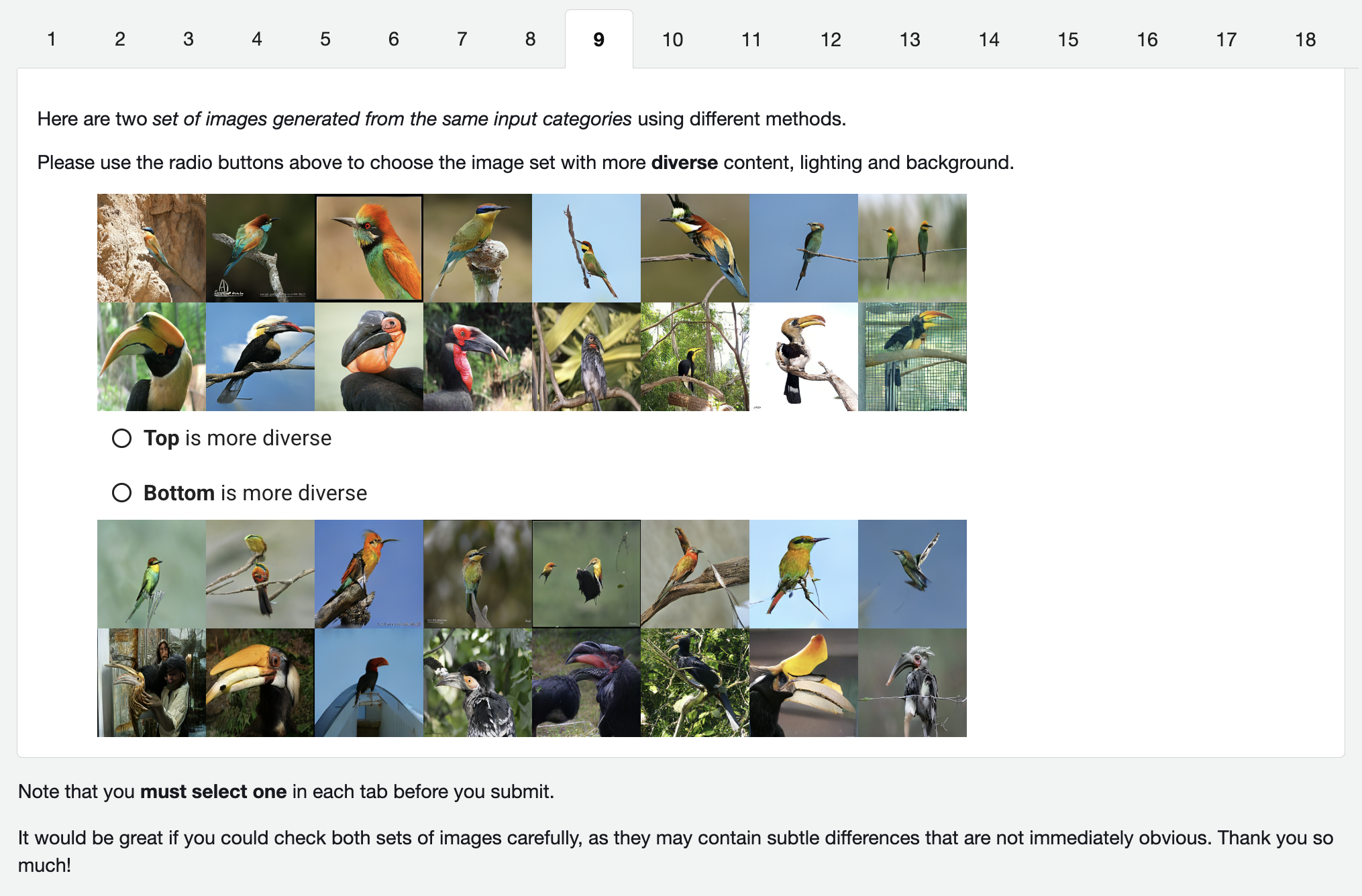}
    \caption{User Interface of diversity evaluation.}
    \label{fig:diversity_eval}
    \vspace{-0.1in}
\end{figure*}

\paragraph{User Preference Study} We present more details about our user preference study. In Figure~\ref{fig:quality_eval}, we provide the interface of quality evaluation, where the users are presented with two generated images of the same class. Given the provided text prompts, the users are asked to select the one with better quality. The interface of diversity evaluation is provided in Figure~\ref{fig:diversity_eval}. Different from evaluating quality, users are shown with 16 random samples from two classes to determine the more diverse groups. 

\paragraph{Inference Efficiency} As the NAR formalism requires much fewer decoding steps than existing AR and DMs, we also study the actual inference speeds to show the efficiency of StraIT. We follow the (18+6)-step allocation in the main paper and conduct inference on TPUv3 chips. Note that in order to consistently compare with previous NAR methods~\cite{maskgit} in terms of parameters, we regard a single step as going through our whole model: including the top and bottom-level transformer. Therefore, this step arrangement is even faster than our marked 12-step inference. However, when comparing with relevant works such as Draft-and-revise~\cite{lee2022draft} or Token-critic~\cite{lezama2022improved}, StraIT can be clarified as using 24 inference steps. On Google TPUv3 and Nvidia V100, it respectively takes 0.18s and 0.39s for StraIT to sample one image, which is order-of-magnitude faster than existing methods such as ADM~\cite{ADM} and LDM~\cite{LDM}.

\begin{figure}[htb]
    \centering
    \includegraphics[width=0.9\columnwidth]{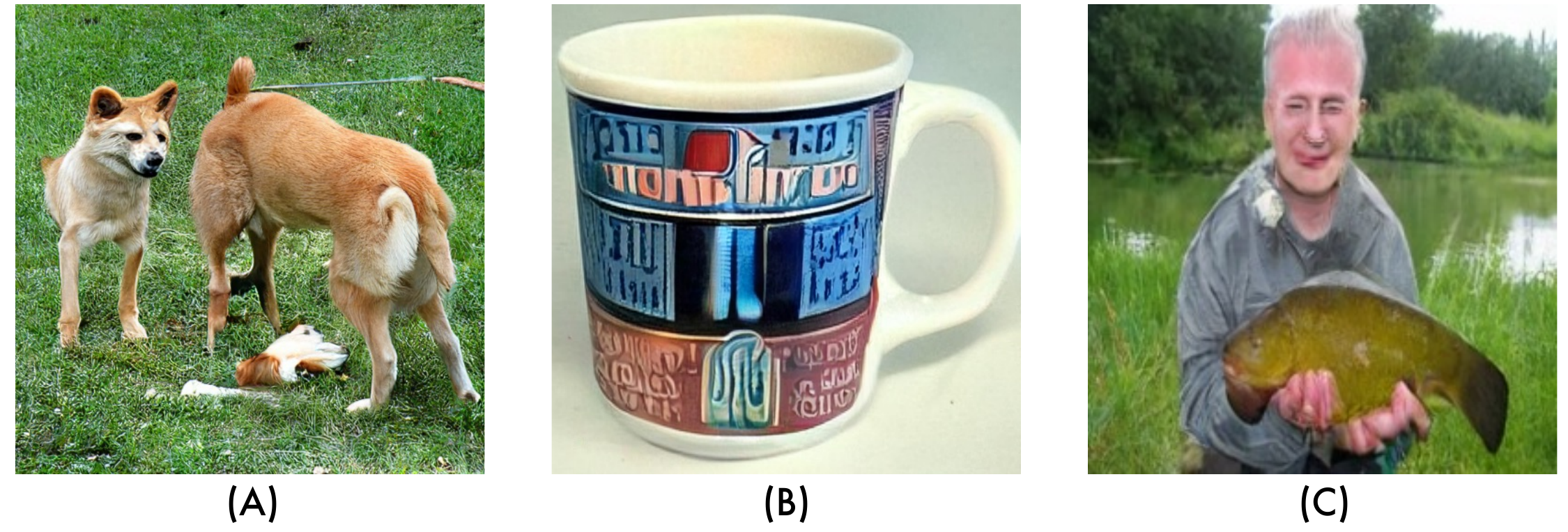}
    \caption{Three generated samples that show limitations, including (A) decoding error; (B) spelling; (C) artifacts on faces and hands.}
    \label{fig:limitation}
    \vspace{-0.1in}
\end{figure}

\suppsection{Limitations and future works}
\label{appendix:limitation}

\paragraph{Limitations} There also exist several limitations in StraIT. In Figure~\ref{fig:limitation}, we show some major limitations of our approach. Existing iterative decoding strategies in NAR still have spaces to improve. In~(A), we demonstrate a failure case from parallel confidence-based decoding. Predicted tokens with high confidence are kept without replacement, leading to  compounding decoding error from early steps. In (B) and (C), we show the model's incapability on spelling and generating tiny faces or hands, which results in undesired artifacts over these complex structures. These issues are also demanding in existing generative models~\cite{parti,LDM,daras2022discovering}. Meanwhile, the quadratic computation and memory costs in self attention makes scaling to higher resolution challenging.

\paragraph{Future Works} With the improved generation quality and clear inference speedups over AR and DMs, StraIT boosts high fidelity NAR generation and opens up many possibility on practical usages. Serving as a general framework, we expect StraIT to facilitate text-to-image~\cite{dalle2,imagen,parti,LDM} and video generation~\cite{villegas2022phenaki}. More importantly, less inference steps enable optimization and deployment much easier. 

Besides these promising directions, the circumstances in limitations also remain future works, including finding better decoding algorithms~\cite{lezama2022improved}, reducing memory costs, and applying cascaded strategies~\cite{CDM} to enhance details. 

\end{document}